%% file: main.tex
\documentclass[runningheads]{llncs}

\usepackage{eccv}

\usepackage{eccvabbrv}

\usepackage{graphicx}
\usepackage{booktabs}

\usepackage[accsupp]{axessibility}  

\usepackage{relsize}

\usepackage[pagebackref,breaklinks,colorlinks,citecolor=eccvblue]{hyperref}

\usepackage{orcidlink}

\usepackage{afterpage}

\input{preamble}

\definecolor{cvprblue}{rgb}{0.21,0.49,0.74}

\usepackage{url}            
\usepackage{booktabs}       
\usepackage{amsfonts}       
\usepackage{nicefrac}       
\usepackage{microtype}      
\usepackage{xcolor}         
\usepackage{graphicx,wrapfig,amsmath}

\usepackage{pifont}
\newcommand{\cmark}{\ding{51}}%
\newcommand{\xmark}{\ding{55}}%

\usepackage{xspace}
\makeatletter
\DeclareRobustCommand\onedot{\futurelet\@let@token\@onedot}
\def\@onedot{\ifx\@let@token.\else.\null\fi\xspace}
\def\eg{\emph{e.g}\onedot} 
\def\ie{\emph{i.e}\onedot}

\def\wrt{w.r.t\onedot} 

\usepackage{makecell}

\usepackage{amssymb}
\usepackage{color,colortbl}

\usepackage{multirow}
\usepackage{pifont}
\usepackage{gensymb}
\usepackage{enumitem}

\definecolor{lightgreen}{RGB}{100,220,100}
\definecolor{darkgreen}{RGB}{30,150,30}
\definecolor{darkblue}{RGB}{0,0,127}
\definecolor{darkyellow}{RGB}{171,133,0}
\definecolor{darkred}{RGB}{180,20,20}
\definecolor{darkmagenta}{RGB}{127,0,127}
\definecolor{darkcyan}{RGB}{0,127,127}
\definecolor{lightred}{RGB}{220,20,20}

\newcommand{\darkgreen}[1]{\textcolor{darkgreen}{#1}}
\newcommand{\lightred}[1]{\textcolor{lightred}{#1}}

\definecolor{purple}{HTML}{9900ff}
\definecolor{darkpink}{HTML}{ff00ff}
\definecolor{maroon}{HTML}{980000}

\definecolor{coolblack}{rgb}{0.0, 0.23, 0.64}

\usepackage{stmaryrd}

\definecolor{darkpastelgreen}{rgb}{0.01, 0.75, 0.24}

\usepackage{lipsum}
\usepackage{bbm}

\newcommand{\expectation}{\mathop{\mathbb{E}}}
\newcommand{\newblue}[1]{#1}

\definecolor{darkpastelgreen}{rgb}{0.01, 0.75, 0.24}

\begin{document}

\title{Interpretability-Guided Test-Time \newblue{Adversarial} Defense} 


\author{Akshay Kulkarni\inst{1}\orcidlink{0000-0003-3379-2238} \and
Tsui-Wei Weng\inst{2}}

\authorrunning{A. Kulkarni and T. W. Weng}

\institute{CSE, UC San Diego \and
HDSI, UC San Diego\\
\email{\{a2kulkarni,lweng\}@ucsd.edu}}

\maketitle

\begin{abstract}

We propose a novel and low-cost test-time \newblue{adversarial} defense by devising interpretability-guided neuron importance ranking methods to identify neurons important to the output classes. Our method is a training-free approach that can significantly improve the robustness-accuracy tradeoff while incurring minimal computational overhead. While being among the most efficient test-time defenses (4$\times$ faster), our method is also robust to a wide range of black-box, white-box, and adaptive attacks that break previous test-time defenses. We demonstrate the efficacy of our method for CIFAR10, CIFAR100, and ImageNet-1k on the standard RobustBench benchmark (with average gains of 2.6\%, 4.9\%, and 2.8\% respectively). We also show improvements (average 1.5\%) over the state-of-the-art test-time defenses even under strong adaptive attacks.

\end{abstract}

\footnotetext[3]{Code: \href{https://github.com/Trustworthy-ML-Lab/Interpretability-Guided-Defense}{https://github.com/Trustworthy-ML-Lab/Interpretability-Guided-Defense}}
\footnotetext[4]{This work is accepted for publication at ECCV 2024.}

\input{1_intro}

\input{2_background}

\input{3_observation}

\input{4_method}

\input{5_experiment}
\input{6_conclusion}

\appendix
\section*{Appendix}

\noindent
In this document, we provide extensive implementation details and additional performance analysis. Towards reproducible research, we will release our complete codebase and saved importance rankings for the reported base models \href{https://github.com/Trustworthy-ML-Lab/Interpretability-Guided-Defense}{in this GitHub repository}.
The appendix is organized as follows:

\renewcommand{\labelitemii}{$\circ$}

\begin{itemize}
\setlength{\itemindent}{-0mm}
    \item Section~\ref{sup:sec:related_work}: Related Work
    \item Section~\ref{sup:sec:implementation}: Implementation details
    \begin{itemize}
        \setlength{\itemindent}{-0mm}
        \item Analysis experiments setup (Sec.\ \ref{sup:subsec:analysis_expt_setup})
        \item Base models (Sec.\ \ref{sup:subsec:base_models})
        \item Evaluation setup (Sec.\ \ref{sup:subsec:evaluation})
        \item Miscellaneous details (Sec.\ \ref{sup:subsec:compute_details})
    \end{itemize}
    \item Section~\ref{sup:sec:expts}: Experiments
    \begin{itemize}
        \setlength{\itemindent}{-0mm}
        \item Extended comparisons (Sec.\ \ref{sup:subsec:extended_comp}, Table \ref{sup:tab:comparison_test_time_ext})
        \item Extended analysis (Sec.\ \ref{sup:subsec:sensitivity}, Table \ref{sup:tab:varying_eps}, \ref{sup:tab:comparison_test_time}, \ref{sup:tab:class_disparity}, Fig.\ \ref{sup:fig:rs_sensitivity}, \ref{sup:fig:loss_surface}, \ref{sup:fig:varying_steps})
        \item Efficiency analysis (Sec.\ \ref{sup:subsec:efficiency}, Table \ref{sup:tab:efficiency})
    \end{itemize}
\end{itemize}{}

\input{suppl/1_related_work}

\input{suppl/2_impl_details}

\input{suppl/3_analysis}


\input{main.bbl}
\end{document}

%% file: preamble.tex
\usepackage[dvipsnames]{xcolor}
\newcommand{\red}[1]{{\color{red}#1}}


%% file: 1_intro.tex
\section{Introduction}

Despite the popularity and success of deep neural networks (DNNs), they have been shown to be brittle against carefully designed small perturbations that are imperceptible to humans \cite{goodfellow2015explaining}. These ``adversarial'' perturbations pose significant risks against deploying DNNs for safety-critical tasks like autonomous driving. To mitigate these risks, there has been great interest in developing defenses, \eg adversarial training algorithms \cite{wong2020fast,zhang2019theoretically,sriramanan2020guided,wu2020adversarial} to obtain a model that is more robust to adversarial samples.

\begin{figure}[t]
    \centering
    \includegraphics[width=0.8\linewidth]{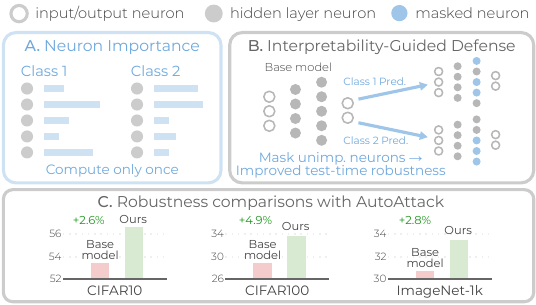}
    \caption{Interpretability-guided masking based on neuron importance ranking for test-time adversarial defense. 
    }
    \label{fig:teaser}
\end{figure}

However, adversarial training is much more expensive than standard training (\eg a strong defense like TRADES \cite{zhang2019theoretically} takes 150-200$\times$ longer than standard training). This inspired training-free defenses known as test-time defenses \cite{kang2021stable,alfarra2022combating,mao2021adversarial,shi2021online}. The underlying principles in existing test-time defenses are either input purification or model adaptation: input purification \cite{alfarra2022combating,shi2021online} modifies the input with an additional perturbation to nullify any adversarial perturbation, whereas model adaptation \cite{kang2021stable,chen2021towards} updates model parameters or adds new parameters at test-time to make the overall model more robust. Despite eliminating the need for training, most existing methods are still quite expensive, 8-500$\times$ more than a standard forward pass \cite{croce2022evaluating}. Moreover, existing test-time defenses have been broken under strong adaptive attacks \cite{croce2022evaluating}. Motivated by these limitations, in this paper, we intend to find an \textit{efficient} and \textit{effective} alternative to improve robustness at test-time, so that the defense does not break under adaptive attacks.

To design an effective defense, we leverage recent neuron-level interpretability methods \cite{bau2017network,oikarinen2023clipdissect}, which allow one to interpret the functionalities of individual neurons. In our analysis, we observed that for successful adversarial examples, the majority of activations at an intermediate layer shift from important neurons for the ground-truth class to the important neurons for other classes. Hence, we propose a novel test-time Interpretability-Guided Defense (\textbf{IG-Defense}) enabled by ``neuron importance ranking'' where class-wise importance of each neuron is computed (Fig.\ \ref{fig:teaser}\red{A}). To design effective neuron importance ranking, we repurpose our analysis experiments and existing neuron-level interpretability works to propose two novel neuron importance ranking methods. The key idea of our \textbf{IG-Defense} is to restrict the observed activation shift by masking unimportant neurons (Fig.\ \ref{fig:teaser}\red{B}). We demonstrate the usefulness of our insights by extensively evaluating our \textbf{IG-Defense} on the standard RobustBench benchmark (Fig.\ \ref{fig:teaser}\red{C}). We observe consistent gains for CIFAR10 (\darkgreen{+2.6\%}), CIFAR100 (\darkgreen{+4.9\%}), and ImageNet-1k (\darkgreen{+2.8\%}). We also devise an ensemble of strong adaptive attacks specifically targeting the weaknesses of \textbf{IG-Defense} and observe consistent gains (\darkgreen{+1.5\%}), unlike existing test-time defenses that break under adaptive attacks. Due to our simple yet grounded approach, \textbf{IG-Defense} is among the most efficient test-time defenses ($4\times$ faster than the strongest existing defense).

\noindent
Our primary contributions can be summarized as:
\begin{itemize}
    \item We are the first to propose a neuron-interpretability-guided test-time defense (\textbf{IG-Defense}) utilizing neuron importance ranking (\texttt{CD-IR}, \texttt{LO-IR}) to improve adversarial robustness. \textbf{IG-Defense} is training-free, efficient, and effective.
    \item We uncover novel insights into improving adversarial robustness by analyzing adversarial attacks through the lens of neuron-level interpretability in Sec.\ \ref{sec:analysis}.
    \item Our proposed \textbf{IG-Defense} consistently improves the robustness on standard CIFAR10, CIFAR100, and ImageNet-1k benchmarks. We also demonstrate improved robustness upto 3.4\%, 3.8\%, and 1.5\% against a wide range of white-box, black-box, and adaptive attacks respectively with the lowest inference time  (4$\times$ faster) among existing test-time defenses.
\end{itemize}

%% file: 2_background.tex
\section{Related Work}

\noindent
\textbf{Test-time adversarial robustness.} 
Most of the test-time defenses can be categorized into two approaches: input purification or model adaptation. In input purification, the adversarial input is converted to a safer input by an additional perturbation. For instance, finding a perturbation that either increases top class confidence \cite{alfarra2022combating}, minimizes contrastive loss \cite{mao2021adversarial}, or minimizes a self-supervised objective \cite{shi2021online}. \cite{hwang2022aidpurifier} crafts the perturbation by fooling a real-adversarial discriminator \cite{hwang2022aidpurifier} and \cite{yoon2021adversarial} uses an input passed through a score-based generative model. For model adaptation \cite{chen2021towards, wu2020rmc, nayak2022dad}, the model parameters are either updated at test-time or new parameters are introduced. However, existing test-time defenses are quite expensive (8-500$\times$ of single forward pass) and break under adaptive attacks. In contrast, our method is fast and effective (1.5\% gain under adaptive attacks with only 2$\times$ inference time), and is the first test-time defense to leverage neuron-level interpretability information.

\noindent
\textbf{Neuron interpretability methods.}
To understand the functionalities (\ie \textit{concepts}) of individual neurons in DNNs, early works like NetDissect \cite{bau2017network, bau2020understanding} compare the neuron activation patterns with patterns of predefined concept label masks. MILAN \cite{hernandez2022natural} produced natural language descriptions using a supervised image captioning model trained on a large-scale concept-labeled dataset. Recently, CLIP-Dissect \cite{oikarinen2023clipdissect} leveraged the CLIP paradigm \cite{radford2021learning} to identify concepts of neurons more efficiently than prior work and without the need of training or densely-concept-labeled datasets. There are other works to identify neuron concepts through clustering \cite{gerasimou2020importance, xie2022npc}, via language models~\cite{bai2024describe}, or using linear combinations~\cite{oikarinen2024linear}. In this work, we focus on repurposing CLIP-Dissect due to its efficiency and simplicity as well as NetDissect to propose neuron importance ranking methods.

\noindent
\textbf{Connections between interpretability and robustness.} 
Prior works attempt to improve robustness via input-dependent interpretability methods like saliency or class activation maps \cite{mangla2020on,boopathy2020proper}.
\newblue{Wu et al.\ \cite{wu2023attention} observed that adversarial attacks lead to changes in attention maps, and proposed a training-time defense with losses to enforce similar attention maps for clean and adversarial samples.} Several works \cite{eigen2021topkconv, bai2021improving, kundu2021tunable, xiao2020enhancing, sehwag2020hydra, zhao2023holistic} use sparsity to improve adversarial robustness. However, they require retraining after or while sparsity constraints are applied. In contrast, we indirectly leverage sparsity without requiring any training. NetDissect \cite{bau2017network} investigated the connection between adversarial attacks and interpretability information from their method, but they did not explore how to improve robustness.
Hence, we aim to study the unexplored connection of neuron-level interpretability and improving adversarial robustness.

%% file: 3_observation.tex
\section{Analysis Experiments}
\label{sec:analysis}

To understand the connection of interpretability of individual neurons with robustness, we aim to design an analysis experiment. One way would be to investigate the changes in neuron activations before and after an adversarial attack. In neuron-labeling works \cite{bau2020understanding, mu2020compositional}, they assign an explainable concept to each neuron (or an activation channel in case of CNNs). Specifically, NetDissect \cite{bau2020understanding} qualitatively show that a targeted adversarial attack causes a drop in peak activations for neurons whose concepts are related to the original class, while causing an increase in peak activations for neurons whose concepts are related to the target class of the attack. They also show that the peak activation change is lower for neurons with unrelated concepts. While they only perform the analysis to showcase their interpretability, we are the first to analyze adversarial attacks through neuron-level interpretability with the explicit goal of obtaining insights into improving adversarial robustness.

For a more thorough analysis, we observe the average activation change (instead of peak change) and associate the neurons with the category names directly instead of concepts. First, peak activation change may be misleading since it may be caused by specific images while average change takes all images into account for a more well-rounded analysis. Second, our automated evaluation eliminates the need for manual association of concepts with categories \cite{bau2020understanding}, which could be difficult, expensive, and noisy, especially with a large number of related categories (\eg ImageNet classes). We analyze both successful and unsuccessful attacks while \cite{bau2020understanding} only focused on successful attacks.

\begin{figure*}[t]
    \centering
    \includegraphics[width=0.8\linewidth]{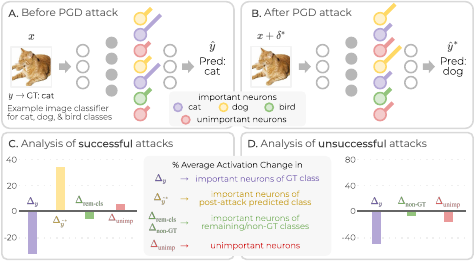}
    \caption{
    \textbf{A.} Cat-dog-bird image classifier before a PGD attack. \textbf{B.} After PGD attack, the prediction changes from the ground truth (GT) cat to dog since the activations of neurons important to dog class increase while those important to cat class decrease. \\
    \textbf{C.} Empirically, successful PGD attacks show a decrease in activations of important GT class neurons while those of post-attack predicted class' important neurons increase. \\
    \textbf{D.} For unsuccessful PGD attacks, the activations of all neurons reduce even though the prediction remains the same as before the attack.
    }
    \label{fig:analysis_expt}
\end{figure*}

Concretely, we find top-$k$ important neurons related to each task category using neuron importance ranking (discussed in Sec.\ \ref{sec:loir}). With this, in Fig.\ \ref{fig:analysis_expt}, we analyze the change in activations corresponding to ground-truth (GT) and non-GT categories before and after an attack. We discuss some preliminaries before delving into further details.

\noindent
\textbf{Preliminaries.}
We denote the dataset as $\mathcal{D}=\{(x, y)\}$ where $x\in\mathcal{X}$ is the input image, $y\in\mathcal{C}$ is the label, $\mathcal{C}$ is the label set, and $|\mathcal{C}|$ is the number of classes. Given a pretrained model $f\!:\!\mathcal{X}\!\to\!\mathcal{C}$ and its activations at the layer being studied $A\!:\!\mathcal{X}\!\to\! \mathbb{R}^{N\times H\times W}$, we can rank the importance of a neuron (out of $N$ neurons/channels) by assessing its contribution to each class. We call this ``neuron importance ranking'' and propose two algorithms for this (Sec.\ \ref{sec:loir}). Then, we can compute the activations of the top-$k$ important neurons for any class $c_i$ as $A_{c_i}(x)\!=\! A(x)_{[n_{c_i}]} \!\in\! \mathbb{R}^{k\times H \times W}$, where $n_{c_i}$ are the indices of the top-$k$ activations for class $c_i$. With access to each $n_{c_i}$, the remaining indices absent from the top-$k$ of any class are $n_u \!=\! [N]\setminus \{\cup_{c_i\in\mathcal{C}}\; n_{c_i}\}$, where $[N] \!=\! \{1, 2, \dots, N\}$ and $\setminus$ indicates set-minus operator. These indices $n_u$ denote unimportant neurons, and corresponding unimportant activations can be computed as $A_u(x) \!=\! A(x)_{[n_u]} \in \mathbb{R}^{(N - k|\mathcal{C}|) \times H \times W}$. Next, we define activation change metrics to support our analysis.

\noindent
\textbf{Metrics.}
For an adversarial attack with $\ell_\infty$-norm bound of $\epsilon$, the perturbation for input $x$ is $\delta^* \!=\! \arg\max_{\Vert \delta\Vert_\infty \leq \epsilon} \;\ell(f(x+\delta), y)$ where $\ell$ is the attack objective. Let $\hat{y}^* = f(x+\delta^*)$ be the post-attack prediction.
The average activation change in the important neurons \wrt any class ${c_i}$ is,
\begin{equation}
    \Delta_{c_i} = \mathop{\mathbb{E}}_{(x, y)\in\mathcal{D}} [A_{c_i}(x+\delta^*) - A_{c_i}(x)]
\end{equation}
Replacing ${c_i}$ with GT class $y$, the average activation change in the important neurons of GT class is $\Delta_y$. Similarly, the average activation change in post-attack predicted class neurons is $\Delta_{\hat{y}^*}$, and that in other non-GT class neurons is $\Delta_\text{non-GT}\!=\!\mathbb{E}_{c_i\neq y, \hat{y}^*}[\Delta_{c_i}]$. 
For unsuccessful attacks, where post-attack prediction $\hat{y}^*$ is the same as $y$, we can compute average activation change in the remaining classes (other than GT) as $\Delta_\text{rem-cls}\!=\!\mathbb{E}_{c_i\neq y}[\Delta_{c_i}]$.
Finally, we can also compute the average activation change in unimportant neurons,
\begin{equation}
    \Delta_\text{unimp}=\mathop{\mathbb{E}}_{(x, y)\in\mathcal{D}} [A_u(x+\delta^*) - A_u(x)]
\end{equation}

We empirically compute the above metrics on CIFAR10 with five base models (ResNet18: GAT \cite{sriramanan2020guided}, NuAT \cite{sriramanan2021towards}, OAAT \cite{addepalli2022scaling}, DAJAT \cite{addepalli2022efficient}, and WideResNet-34-10: TRADES-AWP \cite{wu2020adversarial}) and present the mean values in Fig.\ \ref{fig:analysis_expt}\red{C} and Fig.\ \ref{fig:analysis_expt}\red{D}.
See Appendix \ref{sup:subsec:analysis_expt_setup} for more experimental details and corresponding numerical results.

\noindent
\textbf{Observations.}
In Fig.\ \ref{fig:analysis_expt}\red{C}, we observe $\Delta_y \!<\! 0$ and $\Delta_{\hat{y}^*}\!>\!0$, \ie successful adversarial attacks boost the activations that are important to the post-attack predicted class $\hat{y}^*$ while causing a drop for those of the GT class $y$. Further, from Fig.\ \ref{fig:analysis_expt}\red{C}, the important activations for the remaining classes and unimportant activations are only marginally affected by successful attacks. In Fig.\ \ref{fig:analysis_expt}\red{D}, we observe that unsuccessful attacks cause a drop in the activations of all neurons.

\noindent
\textbf{Remarks.}
During both successful and unsuccessful attacks, the activation magnitude of GT class' important neurons drops. Specifically, successful attacks shift the activation magnitude from GT class' important neurons to non-GT class important neurons.
Hence, we hypothesize that adversarial robustness can be improved if the activation shift to non-GT class important neurons is restricted.

Further, given the availability of a huge number of adversarially pretrained models \cite{croce2021robustbench}, we aim to evaluate our hypothesis in a test-time adversarial defense setting, with a low inference overhead and without any training overhead.

%% file: 4_method.tex
\section{Interpretability-Guided Defense}

In this section, we propose a novel, training-free, test-time Interpretability-Guided Defense (\textbf{IG-Defense}). The core idea of our method is based on interpretability-guided masking, described in Sec.\ \ref{sec:igmasking}. In Sec.\ \ref{sec:loir}, we introduce neuron importance ranking, which is a key enabler in the success of our proposed defense.

\noindent
\textbf{Setup.} 
Given a trained model $f\!:\!\mathcal{X}\to \mathcal{C}$, the goal of test-time defense is to improve the robustness of $f$ while retaining its utility (\ie clean accuracy). 
We refer to $f$ as the ``base model'' in the rest of the paper.

\subsection{Interpretability-guided masking}
\label{sec:igmasking}

\begin{figure*}[t]
    \centering
    \includegraphics[width=\linewidth]{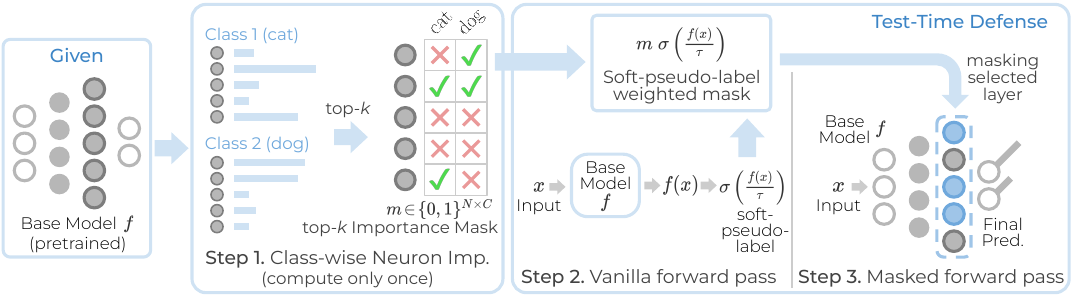}
    \caption{
        \textbf{Step 1.} Given a pretrained base model $f$ (\eg binary classifier here), class-wise neuron importance is computed for a selected layer (Sec.\ \ref{sec:loir}). A top-$k$ mask $m\!\in\!\{0, 1\}^{N\times C}$ is computed to identify top-$k$ neurons important to each class (\eg $k=2, N=5, C=2$ here). \textbf{Step 2.} During evaluation, a soft-pseudo-label $\hat{y}$ is computed using the base model $f$. \textbf{Step 3.} The soft-pseudo-label weighted mask $m\hat{y} = m\; \sigma(\frac{f(x)}{\tau})\in \mathbb{R}^N$ is applied to the selected layer to retain only the important neurons of the pseudo-label class. 
    }
    \label{fig:soft_masking_approach}
\end{figure*}

To validate our hypothesis, we design a simple approach for test-time defense where we mask all the neurons except the important neurons of the correct class. We propose a dual forward pass approach, since access to labels cannot be assumed at test-time. Our approach contains three key steps:
\begin{itemize}
    \item \textbf{Step 1: Neuron Importance Ranking.} First, we compute class-wise neuron importance ranking (Step \red{1} of Fig.\ \ref{fig:soft_masking_approach}). This is computed only once for a given base model $f$. For each layer, we rank every neuron based on its importance to each class. More details are given in Sec.\ \ref{sec:loir}.
    \begin{itemize}
        \item Given the number of important neurons $k$ (hyperparameter), we compute a binary mask $m \!\in\!\{0, 1\}^{N\times C}$ using the indices of important neurons $n_{c_i}\in\mathbb{N}^k$ (from the importance ranking in Sec.\ \ref{sec:loir}) for each class $c_i\in [C]$,
        \begin{equation}
            m{[n_{c_i}, c_i]} = 1\;\forall\;c_i \in [C],\;\;\;\; m{[[N] \setminus n_{c_i}, c_i]} = 0 \;\forall\;c_i \in [C]
        \end{equation}
        \item In the mask $m$, the top-$k$ important neurons of each class $c_i\in [C]$ are retained in $m[:, c_i]$. The remaining neurons unimportant to $c_i$ are masked.
    \end{itemize}
    \item \textbf{Step 2: Vanilla Forward Pass.} In the first forward pass (Step \red{2} of Fig.\ \ref{fig:soft_masking_approach}), we obtain a soft-pseudo-label $\hat{y}=\sigma\left(\frac{f(x)}{\tau}\right)\in (0,1)^C$ where $\sigma(\cdot)$ denotes the softmax function, and $\tau$ denotes the temperature term which controls the sharpness of the soft-pseudo-label.
    \item  \textbf{Step 3: Masked Forward Pass.} In the second forward pass (Step \red{3} of Fig.\ \ref{fig:soft_masking_approach}), we apply a soft-pseudo-label weighted mask \ie $m \hat{y}\in(0, 1)^N$ to the activations of the layer being masked. Let $A:\mathcal{X}\to \mathbb{R}^{N\times H\times W}$ give the activations at the layer being masked, and $h:\mathbb{R}^{N\times H\times W}\to\mathcal{C}$ be the remaining layers after the layer being masked, \ie $f=h\circ A$. Then, the final prediction is given by $\hat{y}'=h(m \hat{y} \cdot A(x))$.
    \begin{itemize}
        \item The masking re-weights the neurons based on $\hat{y}$. However, the temperature term $\tau$ is set to a small value so that $\hat{y}$ contains a sharp probability distribution. Thus, the weighted mask corresponds mainly to a single class and retains only the important neurons of that class.
        \item Also note that this enables gradient flow between the two forward passes, to avoid any gradient masking \cite{athalye2018obfuscated}.
    \end{itemize}
\end{itemize}

Intuitively, activations of a deeper layer (\eg penultimate layer) are masked to minimize any negative impact of the masking on feature extraction, which depends more on the shallower layers \cite{zeiler2014visualizing}. To verify this, we analyze the sensitivity of our method to the layer being masked in Sec.\ \ref{sec:expts:analysis}\red{b}.

\begin{figure*}[t]
    \centering
    \includegraphics[width=\linewidth]{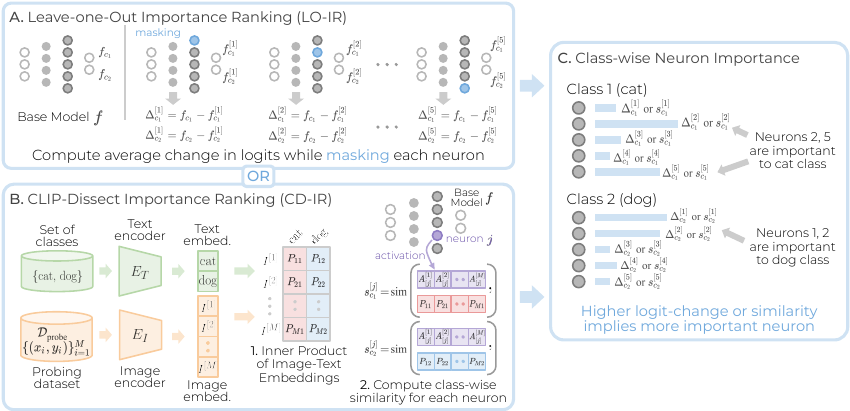}
    \caption{\textbf{A.} Leave-one-Out Importance Ranking (\texttt{LO-IR}) computes class-wise neuron importance as the average change in logits when masking each neuron. \textbf{B.} CLIP-Dissect Importance Ranking (\texttt{CD-IR}) first computes the inner product of class name text embeddings and probing set image embeddings. The importance ranking relies on the similarity between activations of each neuron to the precomputed inner product given the same probing set inputs. \textbf{C.} A higher logit-change $\Delta_{c_i}^{[j]}$ or similarity $s_{c_i}^{[j]}$ implies that neuron $j$ is more important to class $c_i$.}
    \label{fig:cdir}
\end{figure*}

\noindent
\textbf{Dependence on soft-pseudo-label.}
This is an obvious limitation if the soft-pseudo-label is computed directly from the base model $f$. If it is incorrect, masking will yield no improvement in robustness. Hence, an adaptive attack can be designed to ensure that the soft-pseudo-label is incorrect. For instance, a transfer attack that transfers the perturbation from the base model $f$ would yield an incorrect soft-pseudo-label, and the final robustness would be the same as the base model $f$. To circumvent this, we propose to use randomized smoothing in the first forward pass (\ie in Step \red{2} of Fig.\ \ref{fig:soft_masking_approach}).

\noindent
\textbf{Randomized smoothing.}
To obtain the soft-pseudo-label $\hat{y}$ in a more robust manner, we perform randomized smoothing \cite{chen2012ensemble} in the first forward pass.
We add a set of $n_s$ Gaussian noises (with zero mean and standard deviation $\sigma_d$) to each input $x$. For the soft-pseudo-label, the logits of these noisy inputs are averaged and softmax $\sigma$ is applied. Formally,
\begin{equation}
    \hat{y} = 
    \sigma\left(
    \frac{1}{\tau} 
    \expectation_{v_i}
    [f(x+v_i)]
    \right); \;\; v_i \sim \mathcal{N}(0_d, \sigma_d^2 I_d)\; \forall \;i\in [n_s]
\end{equation}
where $0_d, I_d$ are tensors with the same shape as $x$ but containing all zeros and ones respectively. The number of noises sampled $n_s$ and standard deviation $\sigma_d$ are hyperparameters.

\subsection{Importance ranking methods}
\label{sec:loir}

Here, we describe two novel importance ranking methods that can be used in Step \red{1} of our proposed \textbf{IG-Defense}.

\noindent
\textbf{a) Leave-one-Out Importance Ranking} (\texttt{LO-IR}). 
Following the analysis in Sec.\ \ref{sec:analysis}, given a neuron $j$, we can compute its importance to a class $c_i$ as the average change in the class $c_i$ logits, $\Delta_{c_i}^{[j]}$, before and after masking out that particular neuron $j$ (Fig.\ \ref{fig:cdir}\red{A}). Intuitively, a higher logit-change for a particular class implies higher dependence of the network on that neuron, \ie higher importance (Fig.\ \ref{fig:cdir}\red{C}). Formally,
\begin{equation}
    \Delta_{c_i}^{[j]} = \expectation_{(x, y) \in \mathcal{D}_\text{probe} | y=c_i} [f_{c_i}(x) - f_{c_i}^{[j]}(x)] ; \;\; \forall \; c_i\in \mathcal{C}
\end{equation}
where $f^{[j]}$ is the model where $j^\text{th}$ neuron of the selected layer is masked in the base model $f$. And the subscript $c_i$ indicates the logits of class $c_i$.
However, this method could be computationally expensive as average change in logits has to be computed for each neuron. For our proposed method, it is only required once per base model. Thus, it is a feasible option that yields a very good estimate of the importance ranking, which we show in Sec.\ \ref{sec:expts}.

\noindent
\textbf{b) CLIP-Dissect Importance Ranking} (\texttt{CD-IR}).
CLIP-Dissect \cite{oikarinen2023clipdissect} uses the multimodal CLIP model \cite{radford2021learning} to assign concept labels to individual neurons. We extend this idea by replacing concepts with task labels to obtain an importance ranking of neurons for each task label (see Fig.\ \ref{fig:cdir}\red{B}). 

Concretely, probing images are passed through the image encoder $E_I$ to get a set of image embeddings $\{A^{[i]} = E_I(x_i)\}_{i=1}^M$ (shown in orange in Fig.\ \ref{fig:cdir}\red{B}) while the task label names $\mathcal{C}$ are passed through the text encoder $E_T$ to get the corresponding set of text embeddings $\{E_T(c_k)\}_{c_k \in \mathcal{C}}$ (shown in green in Fig.\ \ref{fig:cdir}\red{B}). These embeddings \ie vectors are converted to scalars by taking the inner product of each pair of image-text embeddings, which results in a matrix $P$ with entries $P_{ik}= A^{[i]\top} E_T(c_k)$. Now, for a neuron $j$, a vector of its activations $q_j$ (shown in purple in Fig.\ \ref{fig:cdir}\red{B}) can be obtained where each entry corresponds to each input from the probing dataset. We can compute the similarity score $s_{c_k}^{[j]}$ between each class $c_k$ and neuron $j$ using the $k^{\text{th}}$ column of $P$ and $q_j$. Intuitively, a higher similarity score $s_{c_k}^{[j]}$ implies higher importance of neuron $j$ for class $c_k$ (Fig.\ \ref{fig:cdir}\red{C}). We use the same soft-WPMI similarity metric as the original paper \cite{oikarinen2023clipdissect}. For the probing dataset, we simply use the training dataset for our experiments.

In practice, \texttt{CD-IR} is
computationally less expensive compared to \texttt{LO-IR} \newblue{(\eg 2 mins.\ vs.\ 28 mins.\ for 512 neurons)}. This is because CLIP embeddings can be precomputed and most of the computations can be vectorized easily. As we demonstrate in Sec.\ \ref{sec:expts}, although the importance ranking quality is slightly worse than \texttt{LO-IR} (\ie \texttt{CD-IR} causes a marginal drop in clean accuracy), downstream robustness gains with \texttt{CD-IR} are still significant. Please refer to the Appendix \ref{sup:subsec:efficiency} for details on computational complexity comparisons.

%% file: 5_experiment.tex
\section{Experiments}
\label{sec:expts}

We extensively evaluate our \textbf{IG-Defense} on the standard RobustBench benchmark \cite{croce2021robustbench}. We also follow Croce et al.\ \cite{croce2022evaluating} for properly evaluating test-time defenses and thoroughly test against an ensemble of strong adaptive attacks.

\noindent
\textbf{Experimental setup.} We evaluate our methods on standard datasets, CIFAR10, CIFAR100 \cite{cifar} and ImageNet-1k \cite{imagenet}, with ResNet \cite{he2016deep} and WideResNet \cite{zagoruyko2016wide} base models. We use an $\ell_\infty$ threat model with an $\epsilon$-bound of $8/255$ for CIFAR10/100 and $4/255$ for ImageNet-1k following prior works \cite{croce2021robustbench}. For evaluation, we use the full CIFAR10/100 test set (10000 samples) and the first 5000 ImageNet-1k validation samples, which is the standard setup as per \cite{croce2021robustbench,salman2020adversarially}.
The hyperparameter $k$ for the top-$k$ neurons in \texttt{CD-IR} or \texttt{LO-IR} is mentioned alongside the method name in the results. For randomized smoothing (RS), we use $n_s\!=\!1, \sigma_d\!=\!\frac{\epsilon}{2}$, and $\tau\!=\!0.01$, where $\epsilon$ is the $\ell_\infty$-bound for the attack. We show in Appendix \ref{sup:subsec:sensitivity} that the choices of $n_s\!=\!1, \sigma_d\!=\!\frac{\epsilon}{2}$ are sufficient for our purpose. Unless otherwise mentioned, RS is used and the penultimate layer output is masked. In the penultimate layer, there are 512 neurons (layer4) in ResNet18, 640 neurons (layer3 or block3) in WideResNet-34-10, and 2048 neurons (layer4) in ResNet50.

\noindent
\textbf{Choice of base models.}
We experiment with base models from several single-step defenses like FAT \cite{wong2020fast}, GAT \cite{sriramanan2020guided}, NuAT \cite{sriramanan2021towards}, OAAT \cite{addepalli2022scaling}, and multi-step defenses like TRADES \cite{zhang2019theoretically}, AWP \cite{wu2020adversarial}, DAJAT \cite{addepalli2022efficient}, RTB \cite{salman2020adversarially}, and LAS \cite{jia2022adversarial}. We chose these models to ensure a diverse evaluation and showcase the generality of our \textbf{IG-Defense} across base model defense strategies. We also evaluate with the same base model on different architectures as well as on different datasets for a thorough analysis. Please see Appendix \ref{sup:subsec:base_models} for more details on these defenses.

\input{tabs/cifar10_results}
\input{tabs/in1k_cifar100_results}

\noindent
\textbf{Evaluation.} We evaluate our \textbf{IG-Defense} with the base models against a diverse range of attacks. \textbf{AutoAttack} \cite{croce2020reliable} (AA) is an empirically strong attack with a sequence of white-box AutoPGD (APGD) attacks, fast adaptive boundary (FAB) attack \cite{croce2020minimally}, and black-box Square attack \cite{andriushchenko2020square}. For a stronger adaptive attack, we devise an ensemble of attacks. For each image, the adaptive attack is considered successful if any one attack from the ensemble is successful. With this, we get the image-wise worst-case (\textbf{IW-WC}) robust accuracy. To be adaptive for our \textbf{IG-Defense}, the ensemble also includes transfer attacks that specifically target our weakness, \ie our dependence on pseudo-labels (as discussed in Sec.\ \ref{sec:igmasking}). 

Robust accuracy (IW-WC) is computed with 9 attacks. The first three are computed on the defended model: standard white-box AutoAttack, decision-boundary based black-box RayS attack \cite{chen2020rays}, and white-box APGD with Expectation over Transformation (EoT) \cite{athalye2018obfuscated} attack. For the remaining 6 transfer attacks, we compute perturbations on the base model using 6 attacks: APGD with cross-entropy loss, APGD with Carlini-Wagner loss \cite{carlini2017towards}, targeted APGD attack with difference of logits ratio loss, standard AutoAttack, RayS attack, and APGD+EoT attack.

\noindent
\textbf{Design choices for IW-WC.}
While AutoAttack also uses APGD attacks, it returns unperturbed inputs when the attack is unsuccessful. Following \cite{croce2022evaluating}, the four transfer APGD attacks in IW-WC each return a sample with the highest loss (\ie closest to the decision boundary). This is a stronger attack than AutoAttack for test-time defenses since samples closer to the decision boundary tend to be misclassified under test-time defenses, as observed by \cite{croce2022evaluating}. \textbf{Expectation over Transformation} (EoT) \cite{athalye2018obfuscated} averages gradients over multiple iterations of a particular input to counter the effect of randomness in a defense. Specifically, we use 20 iterations of EoT with APGD (more than $n_s\!=\!1$). While \cite{croce2022evaluating} also recommend using Backward Pass Differentiable Approximation (BPDA) \cite{athalye2018obfuscated} for test-time defenses with non-differentiable components, our approach already includes BPDA since we use a soft-masking approach instead of hard-masking. Refer to Appendix \ref{sup:subsec:evaluation} for complete implementation details.

\subsection{Evaluation on RobustBench} 

\noindent
\textbf{a) CIFAR10.}
In Table \ref{tab:cifar10_rn18}, we evaluate our methods on CIFAR10 with ResNet18 at $\epsilon=8/255$. We observe consistent gains in AutoAttack (avg.\ +2.5\% for \texttt{CD-IR} and +2.55\% for \texttt{LO-IR}) and IW-WC robustness (avg.\ +1.1\% for \texttt{CD-IR} and +1.28\% for \texttt{LO-IR}), with a marginal drop in clean accuracy (avg.\ $-$0.24\% for \texttt{CD-IR} and $-$0.3\% for \texttt{LO-IR}). In Table \ref{tab:cifar10_wrn3410}, we evaluate on CIFAR10 with the more challenging, large WideResNet-34-10 model at $\epsilon=8/255$. We observe consistent gains in IW-WC robustness (avg.\ +1.5\% for \texttt{CD-IR} and +1.7\% for \texttt{LO-IR}), with a marginal drop in clean accuracy (avg.\ $-$0.2\% for \texttt{CD-IR} and $-$0.13\% for \texttt{LO-IR}). 

\input{tabs/test_time_comparisons}

\noindent
\textbf{b) CIFAR100.}
In Table \ref{tab:cifar100_wrn3410}, we evaluate our methods on CIFAR100 with ResNet18 and WideResNet-34-10 at $\epsilon\!=\!8/255$. Despite the increased number of classes, we get consistently improved worst-case (IW-WC) robust accuracy (avg.\ +1.13\% for \texttt{CD-IR} and +1.8\% for \texttt{LO-IR}). However, the clean accuracy drops for \texttt{CD-IR} (avg.\ $-$0.43\% drop), but only marginally for \texttt{LO-IR}. 

\noindent
\textbf{c) ImageNet-1k.}
In Table \ref{tab:in1k_rn50}, we evaluate our methods on ImageNet-1k with ResNet50 at $\epsilon=4/255$. Despite the increased image size and the higher number of classes, we observe consistent gains in worst-case (IW-WC) robustness (avg.\ +0.6\% for \texttt{CD-IR} and +0.85\% for \texttt{LO-IR}) but with a drop in clean accuracy for \texttt{CD-IR} (avg.\ $-$1.47\% similar to CIFAR100). Given the scale of ImageNet-1k, even robustness improvements of $\sim0.5\%$ are quite significant, especially since there is no clean accuracy drop with \texttt{LO-IR}.

\noindent
\textbf{Why does the clean accuracy drop?}
There are two possible reasons for the drop in clean accuracy. First, due to the randomized smoothing, some samples may get incorrect predictions if the random noise is significant enough to change the class prediction \wrt the base model. Second, importance ranking quality and number of neurons masked determine how much clean accuracy can be retained. That is, better quality or lower number of neurons masked would guarantee better clean accuracy retention. However, simply masking lower number of neurons may also reduce the improvements in robustness.

We see in Table \ref{tab:cifar100_wrn3410}, \ref{tab:in1k_rn50} that clean accuracy drop for \texttt{CD-IR} is more for CIFAR100 and ImageNet1k compared to CIFAR10 (Table \ref{tab:cifar10_rn18}, \ref{tab:cifar10_wrn3410}). This is possibly because the ratio of number of neurons to number of classes is decreasing, \ie more number of neurons will be important to multiple classes. This cannot be captured well by \texttt{CD-IR}, leading to a larger drop in clean accuracy. Note that \texttt{LO-IR} is less susceptible to this problem as the importance ranking of each neuron is computed independently, but it is more expensive. This is why we need to retain more number of important neurons for ImageNet ($k\!=\!750$) than CIFAR ($k\!=\!50$).

Note that the clean accuracy drops are acceptable since they are significantly smaller than the worst-case robustness gains (\eg upto $-$0.6\% and +1.7\% respectively for CIFAR10 in Table \ref{tab:cifar10_rn18}-\ref{tab:cifar10_wrn3410}). Further, in some cases like ImageNet-1k with \texttt{LO-IR} (Table \ref{tab:in1k_rn50}), there are no drops with pure gains in robustness.

\subsection{Comparisons with existing test-time defenses}

We compare with three recent state-of-the-art test-time defense methods on CIFAR10 in Table \ref{tab:comparison_test_time}. CAAA \cite{alfarra2022combating} (input purification) optimizes an ``anti-adversary'' perturbation in a direction opposite to an adversarial attack, by minimizing the loss based on a pseudo-label, to counter the effect of any adversarial perturbation. HD or Hedge Defense \cite{wu2021attacking} (input purification) finds a perturbation that maximizes the cross-entropy loss summed over all classes. SODEF \cite{kang2021stable} (model adaptation) trains a neural ODE block with Lyapunov stability constraints between the pretrained feature extractor and classifier, which reduces the influence of the adversarial perturbation.
In contrast, our method uses neither input purification nor model adaptation, since we only mask the activations. 

In Table \ref{tab:comparison_test_time}, we observe that existing test-time defenses like HD, CAAA, and SODEF do not significantly improve, but rather also reduce the worst-case robust accuracy (IW-WC) compared to the base model. Based on Croce et al.\ \cite{croce2022evaluating}, CAAA is the strongest test-time defense among the nine defenses that they evaluated, since its worst-case robust accuracy did not drop below that of the base model. Hence, we only compare with the two strongest input purification defenses, CAAA and HD as well as an efficient model adaptation defense, SODEF from the nine defenses in \cite{croce2022evaluating}. Overall, we find that our proposed \textbf{IG-Defense} obtains consistent improvements in image-wise worst-case (IW-WC) robust accuracy, unlike existing test-time defenses. Note that we focus mainly on the worst-case (IW-WC) robust accuracy because gains under weaker attacks would not reflect fundamental gains in robustness \cite{croce2022evaluating}.

\begin{figure*}[t]
    \centering
    \includegraphics[width=1.0\linewidth]{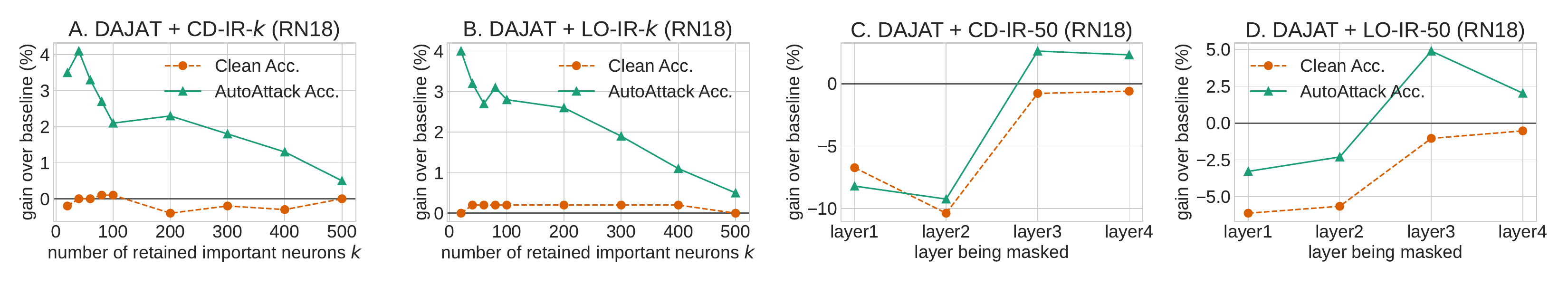}
    \caption{ 
        Sensitivity analyses for number of retained important neurons $k$ and layer being masked with CIFAR10. For \textbf{A}, \textbf{B}: layer4 with total 512 neurons is used. A lower $k$ and a deeper layer yield better adversarial robustness with minimal loss of clean accuracy.
    }
    \label{fig:sensitivity_analyses}
\end{figure*}

\subsection{Analysis}
\label{sec:expts:analysis}

\noindent
\textbf{a) Ablation study.} In Table \ref{tab:ablations}, we study the significance of each proposed component of our approach. First, we start with the baseline (row \#1) which uses only one forward pass without masking and without randomized smoothing (RS). Next, we add RS to this baseline (row \#2) and observe no gains in robustness and a small drop in clean accuracy. This shows that RS alone cannot improve the robustness. Then, we perform two forward passes without masking and without RS (row \#3). Here, the robust accuracy is the same as the baseline indicating that there is no gradient masking with two forward passes. Next, we perform random masking with two forward passes, without and with RS (row \#4, \#5). Each class is assigned a random set of neurons as the ``important'' neurons to perform the masking. We observe no gains in robustness, validating the significance of our neuron importance ranking. Finally, we show the impact of RS with our \texttt{CD-IR} (row \#6, \#7) and \texttt{LO-IR} based masking (row \#8, \#9). Here, we observe consistent gains in robustness, with a small drop in clean accuracy since RS likely changed the prediction \wrt the base model for a few samples.

\input{tabs/ablations}

\noindent
\textbf{b) Sensitivity analyses.} 
We study the sensitivity of our proposed methods to the number of retained important neurons $k$ in Fig.\ \ref{fig:sensitivity_analyses}\red{A}, \red{B} on CIFAR10. Specifically, we test it on ResNet18 layer4 which has total 512 neurons and we vary $k$ from 10 to 500. We observe that the robustness improvement is higher at lower $k$, although lower than 10 may reduce the clean accuracy due to a high loss of information. For \texttt{CD-IR}, the clean accuracy can be retained up to $k=100$ while for \texttt{LO-IR}, it can be retained up to at least $k=400$. As explained earlier, it is due to the relatively lower importance ranking quality of \texttt{CD-IR} \wrt \texttt{LO-IR}. However, both methods are fairly insensitive to the choice of $k$.

We also study the sensitivity to choice of layer being masked in Fig.\ \ref{fig:sensitivity_analyses}\red{C}, \red{D} on CIFAR10 with ResNet18. We fix the number of neurons being retained $k$ to 50 and vary the layer being masked from layer1 (shallower) to layer4 (deeper). We observe that masking deeper layers is better, since masking shallower layers affects the feature extraction \cite{zeiler2014visualizing}, leading to a large drop in clean accuracy, and consequently a drop in robust accuracy. For all the other experiments, we simply mask the penultimate layer since the clean accuracy drop is the lowest.
\newblue{Please refer to Appendix \ref{sup:sec:expts} for more analysis experiments.}

\noindent
\textbf{c) Inference time comparisons.}
Since we use $n_s=1$ in randomized smoothing and perform two forward passes, the inference time of \textbf{IG-Defense} is $2\times$ of a single forward pass. The compared test-time defenses HD \cite{wu2021attacking}, SODEF \cite{kang2021stable}, and CAAA \cite{alfarra2022combating} have computational complexities of $46\times$, $2\times$, and $8\times$ respectively. Also, the recent review paper \cite{croce2022evaluating} shows that the lowest inference time is $2\times$ for existing test-time defenses. Hence, as per Table \ref{tab:comparison_test_time}, \textbf{we achieve the most gains in worst-case robustness with the lowest inference time}.

%% file: tabs/cifar10_results.tex
\begin{table*}[t]
    \centering
    \setlength{\tabcolsep}{3pt}
    \caption{Evaluating our \textbf{IG-Defense} (\texttt{CD-IR}, \texttt{LO-IR}) on \textbf{CIFAR10} with \textbf{ResNet18} and $\ell_\infty, \epsilon=8/255$. IW-WC indicates image-wise worst-case, AA indicates AutoAttack.}
    \resizebox{\textwidth}{!}{
\begin{tabular}{@{}lcclcclccl@{}}
\toprule
\multicolumn{1}{r}{\textbf{Base Model:}}                             & \multicolumn{3}{c}{NuAT \cite{sriramanan2021towards} (RN18)}                                                                                                                                              & \multicolumn{3}{c}{OAAT \cite{addepalli2022scaling} (RN18)}                                                                                                                                              & \multicolumn{3}{c}{DAJAT \cite{addepalli2022efficient} (RN18)}                                                                                                                                             \\
\cmidrule(lr){2-4} \cmidrule(lr){5-7} \cmidrule(lr){8-10}
{\begin{tabular}[c]{@{}l@{}}Test-Time\\ Defense\end{tabular}} & \begin{tabular}[c]{@{}c@{}}Clean\\ Acc.\end{tabular} & \begin{tabular}[c]{@{}c@{}}Robust Acc.\\ (AA)\end{tabular} & \begin{tabular}[c]{@{}c@{}}Robust Acc.\\ (IW-WC)\end{tabular} & \begin{tabular}[c]{@{}c@{}}Clean\\ Acc.\end{tabular} & \begin{tabular}[c]{@{}c@{}}Robust Acc.\\ (AA)\end{tabular} & \begin{tabular}[c]{@{}c@{}}Robust Acc.\\ (IW-WC)\end{tabular} & \begin{tabular}[c]{@{}c@{}}Clean\\ Acc.\end{tabular} & \begin{tabular}[c]{@{}c@{}}Robust Acc.\\ (AA)\end{tabular} & \begin{tabular}[c]{@{}c@{}}Robust Acc.\\ (IW-WC)\end{tabular} \\ \midrule
None                                                                 & 82.21                                                & 50.58                                                 & 50.54                                                         & \textbf{80.23}                                       & 51.10                                                 & 51.01                                                         & 85.71                                                & 52.50                                                 & 52.45                                                         \\
\texttt{CD-IR}-50 (\textit{Ours})                                                             & \textbf{82.46}                                       & 52.08                                                 & 51.58 \darkgreen{\smaller(+1.04)}                                                 & 79.85                                                & 54.79                                                 & \textbf{52.35 \darkgreen{\smaller(+1.34)}}                                        & 85.11                                                & \textbf{54.81}                                                 & \textbf{53.36 \darkgreen{\smaller(+0.91)}}                                        \\
\texttt{LO-IR}-50 (\textit{Ours})                                                             & 82.17                                                & \textbf{52.41}                                        & \textbf{52.19 \darkgreen{\smaller(+1.65)}}                                        & 79.90                                                & \textbf{54.90}                                        & 52.30 \darkgreen{\smaller(+1.29)}                                                 & 85.18                                                & 54.53                                                 & 53.34 \darkgreen{\smaller(+0.89)} \\                         \bottomrule            
\end{tabular}
}
\label{tab:cifar10_rn18}
\end{table*}

\begin{table*}[t]
    \centering
    \setlength{\tabcolsep}{3pt}
    \caption{Evaluating our \textbf{IG-Defense} (\texttt{CD-IR}, \texttt{LO-IR}) on \textbf{CIFAR10} with \textbf{WRN-34-10} and  $\ell_\infty, \epsilon=8/255$. IW-WC indicates image-wise worst-case, AA indicates AutoAttack.}
    \resizebox{\textwidth}{!}{
\begin{tabular}{@{}lcclcclccl@{}}
\toprule
\multicolumn{1}{r}{\textbf{Base Model:}}                    & \multicolumn{3}{c}{NuAT \cite{sriramanan2021towards} (WRN-34-10)}                                                                                                                                         & \multicolumn{3}{c}{TR-AWP \cite{zhang2019theoretically} (WRN-34-10)}                                                                                                                                       & \multicolumn{3}{c}{DAJAT \cite{addepalli2022efficient} (WRN-34-10)}                                                                                                                                        \\
\cmidrule(lr){2-4} \cmidrule(lr){5-7} \cmidrule(lr){8-10}
{\begin{tabular}[c]{@{}l@{}}Test-Time\\ Defense\end{tabular}} & \begin{tabular}[c]{@{}c@{}}Clean\\ Acc.\end{tabular} & \begin{tabular}[c]{@{}c@{}}Robust Acc.\\ (AA)\end{tabular} & \begin{tabular}[c]{@{}c@{}}Robust Acc.\\ (IW-WC)\end{tabular} & \begin{tabular}[c]{@{}c@{}}Clean\\ Acc.\end{tabular} & \begin{tabular}[c]{@{}c@{}}Robust Acc.\\ (AA)\end{tabular} & \begin{tabular}[c]{@{}c@{}}Robust Acc.\\ (IW-WC)\end{tabular} & \begin{tabular}[c]{@{}c@{}}Clean\\ Acc.\end{tabular} & \begin{tabular}[c]{@{}c@{}}Robust Acc.\\ (AA)\end{tabular} & \begin{tabular}[c]{@{}c@{}}Robust Acc.\\ (IW-WC)\end{tabular} \\ \midrule
None                                                        & \textbf{86.32}                                       & 54.75                                                 & 54.72                                                         & \textbf{85.36}                                       & 56.17                                                 & 56.12                                                         & \textbf{88.71}                                       & 57.81                                                 & 57.72                                                         \\
\texttt{CD-IR}-50 (\textit{Ours})                                                    & 86.28                                                & \textbf{57.69}                                        & 56.38 \darkgreen{\smaller(+1.66)}                                                 & 84.98                                                & \textbf{60.26}                                        & 57.46 \darkgreen{\smaller(+1.34)}                                                 & 88.51                                                & \textbf{60.70}                                        & 59.22 \darkgreen{\smaller(+1.50)}                                                 \\
\texttt{LO-IR}-50 (\textit{Ours})                                                    & 86.21                                                & 57.04                                                 & \textbf{56.40 \darkgreen{\smaller(+1.68)}}                                        & 85.17                                                & 59.61                                                 & \textbf{57.88 \darkgreen{\smaller(+1.76)}}                                        & 88.61                                                & 60.08                                                 & \textbf{59.41 \darkgreen{\smaller(+1.69)}} \\
\bottomrule
\end{tabular}
}
\label{tab:cifar10_wrn3410}
\end{table*}

%% file: tabs/in1k_cifar100_results.tex
\begin{table*}[t]
    \centering
    \setlength{\tabcolsep}{3pt}
    \caption{Evaluating our \textbf{IG-Defense} (\texttt{CD-IR}, \texttt{LO-IR}) on \textbf{CIFAR100} with $\ell_\infty, \epsilon\!=\!8/255$. IW-WC indicates image-wise worst-case, AA indicates AutoAttack.}
    \resizebox{\textwidth}{!}{
\begin{tabular}{@{}lcclcclccl@{}}
\toprule
\multicolumn{1}{r}{\textbf{Base Model:}}                    & \multicolumn{3}{c}{LAS \cite{jia2022adversarial} (WRN-34-10)}                                                                                                                                                              & \multicolumn{3}{c}{OAAT \cite{addepalli2022scaling} (RN18)}                                                                                                                                                                  & \multicolumn{3}{c}{DAJAT \cite{addepalli2022efficient} (RN18)}                                                                                                                                                                 \\
\cmidrule(lr){2-4} \cmidrule(lr){5-7} \cmidrule(lr){8-10}
\begin{tabular}[c]{@{}l@{}}Test-Time\\ Defense\end{tabular} & \begin{tabular}[c]{@{}c@{}}Clean\\ Acc.\end{tabular} & \begin{tabular}[c]{@{}c@{}}Robust Acc.\\ (AA)\end{tabular} & \multicolumn{1}{c}{\begin{tabular}[c]{@{}c@{}}Robust Acc.\\ (IW-WC)\end{tabular}} & \begin{tabular}[c]{@{}c@{}}Clean\\ Acc.\end{tabular} & \begin{tabular}[c]{@{}c@{}}Robust Acc.\\ (AA)\end{tabular} & \multicolumn{1}{c}{\begin{tabular}[c]{@{}c@{}}Robust Acc.\\ (IW-WC)\end{tabular}} & \begin{tabular}[c]{@{}c@{}}Clean\\ Acc.\end{tabular} & \begin{tabular}[c]{@{}c@{}}Robust Acc.\\ (AA)\end{tabular} & \multicolumn{1}{c}{\begin{tabular}[c]{@{}c@{}}Robust Acc.\\ (IW-WC)\end{tabular}} \\
\midrule
None                                                        & \textbf{64.89}                                       & 30.74                                                 & 30.71                                                                             & 62.02                                                & 27.15                                                 & 27.12                                                                             & \textbf{65.45}                                       & 27.67                                                 & 27.64                                                                             \\
\texttt{CD-IR}-50 (\textit{Ours})                                                   & 62.13                                                & 33.57                                                 & 31.41 \darkgreen{\smaller(+0.70)}                                                                     & 61.56                                                & 29.51                                                 & 29.19 \darkgreen{\smaller(+2.07)}                                                                     & 64.97                                                & 30.22                                                 & 28.27 \darkgreen{\smaller(+0.63)}                                                                     \\
\texttt{LO-IR}-50 (\textit{Ours})                                                    & 64.85                                                & \textbf{35.69}                                        & \textbf{32.22 \darkgreen{\smaller(+1.51)}}                                                            & \textbf{62.04}                                       & \textbf{32.06}                                        & \textbf{30.12 \darkgreen{\smaller(+3.00)}}                                                            & 65.37                                                & \textbf{32.55}                                        & \textbf{28.54 \darkgreen{\smaller(+0.90)}} \\       
\bottomrule
\end{tabular}
}
\label{tab:cifar100_wrn3410}
\end{table*}

\begin{table*}[t]
    \centering
    \setlength{\tabcolsep}{8pt}
    \caption{Evaluating our \textbf{IG-Defense} (\texttt{CD-IR}, \texttt{LO-IR}) on \textbf{ImageNet-1k} with ResNet50 and $\ell_\infty, \epsilon\!=\!4/255$. IW-WC indicates image-wise worst-case, AA indicates AutoAttack.}
    \resizebox{\textwidth}{!}{
\begin{tabular}{@{}lcllcll@{}}
\toprule
\multicolumn{1}{r}{\textbf{Base Model:}}                    & \multicolumn{3}{c}{FAT \cite{wong2020fast} (RN50)}                                                                                                                                                                                       & \multicolumn{3}{c}{RTB \cite{salman2020adversarially} (RN50)}                                                                                                                                                                                       \\
\cmidrule(lr){2-4} \cmidrule(lr){5-7}
\begin{tabular}[c]{@{}l@{}}Test-Time\\ Defense\end{tabular} & \begin{tabular}[c]{@{}c@{}}Clean\\ Acc.\end{tabular} & \multicolumn{1}{c}{\begin{tabular}[c]{@{}c@{}}Robust Acc.\\ (AA)\end{tabular}} & \multicolumn{1}{c}{\begin{tabular}[c]{@{}c@{}}Robust Acc.\\ (IW-WC)\end{tabular}} & \begin{tabular}[c]{@{}c@{}}Clean\\ Acc.\end{tabular} & \multicolumn{1}{c}{\begin{tabular}[c]{@{}c@{}}Robust Acc.\\ (AA)\end{tabular}} & \multicolumn{1}{c}{\begin{tabular}[c]{@{}c@{}}Robust Acc.\\ (IW-WC)\end{tabular}} \\
\midrule
None                                                        & 55.64                                                & 26.24                                                                     & 26.24                                                                             & 64.10                                                & 34.62                                                                     & 34.62                                                                             \\
\texttt{CD-IR}-750 {\smaller(\textit{Ours})}                                                    & 54.30                                                & 27.28 \darkgreen{\smaller(+1.04)}                                                             & 26.80 \darkgreen{\smaller(+0.56)}                                                                     & 62.50                                                & 36.42 \darkgreen{\smaller(+1.80)}                                                             & 35.30 \darkgreen{\smaller(+0.68)}                                                                     \\
\texttt{LO-IR}-750 {\smaller(\textit{Ours})}                                                    & \textbf{55.88}                                       & \textbf{29.14 \darkgreen{\smaller(+2.90)}}                                                    & \textbf{27.08 \darkgreen{\smaller(+0.84)}}                                                            & \textbf{64.10}                                       & \textbf{37.36 \darkgreen{\smaller(+2.74)}}                                                    & \textbf{35.48 \darkgreen{\smaller(+0.86)}} \\
\bottomrule
\end{tabular}
}
\label{tab:in1k_rn50}
\end{table*}

%% file: tabs/test_time_comparisons.tex
\begin{table*}[t]
    \centering
    \setlength{\tabcolsep}{5pt}
    \caption{Comparison against state-of-the-art test-time defenses with ResNet18 and WideResNet-34-10 on CIFAR10. The number in \darkgreen{(green)} shows the worst-case robustness gain over the base model. The first row is the base model without any test-time defense. Inference time is a multiple of the time for a single forward pass of the base model.}
    \resizebox{\textwidth}{!}{
        \begin{tabular}{lclclclc}
            \toprule
            \multicolumn{1}{r}{\textbf{Base Model $\to$}}  & \multicolumn{2}{c}{OAAT \cite{addepalli2022scaling} (RN18)} & \multicolumn{2}{c}{DAJAT \cite{addepalli2022efficient} (RN18)} & \multicolumn{2}{c}{TR-AWP \cite{wu2020adversarial} (WRN)} & \multirow{3}{*}{\makecell{Inference \\ Time}} \\ \cmidrule(lr){2-3}\cmidrule(lr){4-5}\cmidrule(lr){6-7}
            Test-Time Defense & \begin{tabular}[c]{@{}c@{}}Clean\\ Acc.\end{tabular}          & \begin{tabular}[c]{@{}c@{}}Robust Acc.\\ (IW-WC)\end{tabular}          & \begin{tabular}[c]{@{}c@{}}Clean\\ Acc.\end{tabular}           & \begin{tabular}[c]{@{}c@{}}Robust Acc.\\ (IW-WC)\end{tabular}          & \begin{tabular}[c]{@{}c@{}}Clean\\ Acc.\end{tabular}           & \begin{tabular}[c]{@{}c@{}}Robust Acc.\\ (IW-WC)\end{tabular}          \\
            \midrule
            No test-time def.\ & 80.23 & 51.01 & 85.71 & 52.45 & 85.36 & 56.12 & $1\times$ \\ \midrule
            \textbf{Existing Defenses} \\
            HD \cite{wu2021attacking} & 79.89 & 50.68 {\smaller\lightred{(-0.33)}} & 84.53 & 52.55 {\smaller\darkgreen{(+0.10)}} & 84.78 & 55.72 {\smaller\lightred{(-0.40)}} & $46\times$ \\
            SODEF \cite{kang2021stable} & 80.23 & 50.67 {\smaller\lightred{(-0.34)}} & 84.86 & 52.95 {\smaller\darkgreen{(+0.50)}} & 85.25 & 56.05 {\smaller\lightred{(-0.07)}} & $2\times$ \\
            CAAA \cite{alfarra2022combating} & 80.23 & 51.00 {\smaller\lightred{(-0.01)}} & 85.71 & 52.45 {\smaller\lightred{(+0.00)}} & 85.35 & 56.11  {\smaller\lightred{(-0.01)}} & $8\times$ \\ \midrule
            \textbf{Our IG-Defense} \\
            \texttt{CD-IR}-50  & 79.85 & \textbf{52.35 {\smaller\darkgreen{(+1.34)}}} & 85.11 & \textbf{53.36 {\smaller\darkgreen{(+0.91)}}} & 84.98 & 57.46 {\smaller\darkgreen{(+1.34)}} & \textbf{2}$\times$ \\
            \texttt{LO-IR}-50 & 79.90 & 52.30 {\smaller\darkgreen{(+1.29)}} & 85.18 & 53.34 {\smaller\darkgreen{(+0.89)}} & 85.17 & \textbf{57.88 {\smaller\darkgreen{(+1.76)}}} & \textbf{2}$\times$ \\
            \bottomrule
        \end{tabular}%
    }
    \label{tab:comparison_test_time}
\end{table*}

%% file: tabs/ablations.tex
\begin{wraptable}[16]{R}{0.65\textwidth}
    \centering
    \setlength{\tabcolsep}{4pt}
    \caption{Ablation study of proposed components with TRADES-AWP \cite{wu2020adversarial} (WideResNet-34-10) as the base model. RS indicates use of randomized smoothing in the first forward pass. The first row is the base model performance without any test-time defense.
    }
    \label{tab:ablations}
    \resizebox{0.65\textwidth}{!}{%
        \begin{tabular}{ccccccc}
            \toprule
            \begin{tabular}[c]{@{}c@{}}Row\\ \# \end{tabular} & \begin{tabular}[c]{@{}c@{}}No. of forw.\ \\ passes\end{tabular} & \begin{tabular}[c]{@{}c@{}}Masking\\ type\end{tabular} & RS & \begin{tabular}[c]{@{}c@{}}Clean\\ Acc. \end{tabular} & \begin{tabular}[c]{@{}c@{}}Rob.\ Acc.\ \\(AA)\end{tabular} & \begin{tabular}[c]{@{}c@{}}Rob.\ Acc.\ \\ (IW-WC) \end{tabular} \\ 
            \midrule
            1 & 1 & None & \lightred{\xmark} & \textbf{85.36} & 56.17 & 56.12 \\
            2 & 1 & None & \darkgreen{\cmark} & 84.93 & 56.19 & 56.14 \\
            3 & 2 & None & \lightred{\xmark} & 85.35 & 56.15 & 56.12 \\
            4 & 2 & \texttt{Random}-50 & \lightred{\xmark} & 85.23 & 56.14 & 56.12 \\
            5 & 2 & \texttt{Random}-50 & \darkgreen{\cmark} & 85.33 & 56.20 & 56.15 \\
            6 & 2 & \texttt{CD-IR}-50 & \lightred{\xmark} & 85.09 & 59.00 & 56.12 \\
            7 & 2 & \texttt{CD-IR}-50 & \darkgreen{\cmark} & 84.98 & \textbf{60.26} & 57.46 \\
            8 & 2 & \texttt{LO-IR}-50 & \lightred{\xmark} & 85.33 & 58.59 & 56.12 \\
            9 & 2 & \texttt{LO-IR}-50 & \darkgreen{\cmark} & 85.17 & 59.61 & \textbf{57.88} \\
            \bottomrule
        \end{tabular} 
    }
\end{wraptable}

%% file: 6_conclusion.tex
\section{Conclusion}
In this work, we proposed a novel, training-free, and effective test-time defense (\textbf{IG-Defense}) guided by interpretability through neuron importance ranking. Our \textbf{IG-Defense} outperformed existing test-time defenses on worst-case adaptive attacks while being among the most efficient test-time defenses to date, establishing the new state-of-the-art. We also demonstrated consistent improvements across standard CIFAR and ImageNet-1k benchmarks.

\section*{Acknowledgements}
This work is supported in part by National Science Foundation (NSF) awards CNS-1730158, ACI-1540112, ACI1541349, OAC-1826967, OAC-2112167, CNS-2100237,
CNS-2120019, the University of California Office of the
President, and the University of California San Diego’s California Institute for Telecommunications and Information
Technology/Qualcomm Institute. Thanks to CENIC for the
100Gbps networks. The authors thank the anonymous reviewers for valuable feedback on the manuscript. The authors are partially supported by National Science Foundation awards CCF-2107189, IIS-2313105, IIS-2430539. T.-W. Weng also thanks the Hellman Fellowship for providing research support.

%% file: suppl/1_related_work.tex
\section{Related Work}
\label{sup:sec:related_work}

\textbf{Connections between interpretability and robustness.} 

\noindent
Several works \cite{eigen2021topkconv, bai2021improving, kundu2021tunable, xiao2020enhancing, sehwag2020hydra, zhao2023holistic} use sparsity to improve adversarial robustness. 
Eigen and Sadovnik \cite{eigen2021topkconv} replace standard convolutions with top-$k$ convolutions that only output the sum of top-$k$ channels instead of all channels for each convolutional filter. This requires the model to be trained from scratch.
Bai et al.\ \cite{bai2021improving} introduce learnable channelwise activation suppressing modules which select some channels to be masked. This adds new learnable parameters and also requires the model to be trained from scratch.

A line of work \cite{kundu2021tunable,sehwag2020hydra,zhao2023holistic} focuses on pruning aiming to train a model with robust pruning objectives using both original and adversarial images to obtain a sparser, more robust model.
Xiao et al.\ \cite{xiao2020enhancing} propose a method similar to \cite{eigen2021topkconv} except that they replace ReLU with a top-$k$ activation function instead of modifying the convolutional layers. But their approach deliberately relies on gradient masking, which is apparent from their highly irregular or jagged loss surfaces.
However, these methods require retraining or training from scratch, which is expensive. In contrast, our method indirectly leverages sparsity without requiring any training.

%% file: suppl/2_impl_details.tex
\section{Implementation Details}
\label{sup:sec:implementation}

\subsection{Analysis experiments setup}
\label{sup:subsec:analysis_expt_setup}

For the analysis experiments in Sec.\ \ref{sec:analysis} of the main paper, we use 50-step PGD attack \cite{madry2018towards} on five base models, ResNet18: GAT \cite{sriramanan2020guided}, NuAT \cite{sriramanan2021towards}, OAAT \cite{addepalli2022scaling}, DAJAT \cite{addepalli2022efficient}, and WideResNet-34-10: TRADES-AWP \cite{wu2020adversarial} for the CIFAR10 dataset. 
We provide the results from Fig.\ \ref{fig:analysis_expt} of the main paper in a numerical form in Table \ref{supp:tab:analysis_expt_repeat} for clarity.
The \% activation change is the relative change \wrt the original activation value.

\input{suppl/tabs/analysis_expt}

\subsection{Base models}
\label{sup:subsec:base_models}
We experiment with several single-step (first four) and multi-step defenses (remaining) to demonstrate the wide applicability of our \textbf{IG-Defense}. We briefly discuss the details of the base models' training strategies below.
\begin{itemize}
    \itemsep0em
    \item FAT \cite{wong2020fast} - This uses the randomized fast gradient signed method (R-FGSM) attack during adversarial training.
    \item GAT \cite{sriramanan2020guided} - This uses the maximum margin loss instead of cross-entropy for the attack during adversarial training.
    \item NuAT \cite{sriramanan2021towards} - This uses R-FGSM attack with a nuclear norm regularizer to locally smoothen the loss surface and reduce gradient masking.
    \item OAAT \cite{addepalli2022scaling} - This uses mixup between an original and a perceptually-visible adversarial image (generated with $\epsilon\!=\!24/255$ and LPIPS-based attack) to improve robustness.
    \item DAJAT \cite{addepalli2022efficient} - This uses strong and light data augmentations to improve adversarial training by using separate batch-norm parameters for the two augmentation types.
    \item TRADES \cite{zhang2019theoretically} - This uses a regularizer to push the decision boundary away from data points, which improves robustness.
    \item AWP \cite{wu2020adversarial} - This regularizes the loss landscape flatness by perturbing both input and weights.
    \item RTB \cite{salman2020adversarially} - Adversarial training via multi-step PGD attack.
    \item LAS \cite{jia2022adversarial} - This uses a learnable attack strategy to generate strong attacks for training.
\end{itemize}
We evaluate base models from these methods on the datasets for which they have released their pretrained models.

\subsection{Evaluation Setup}
\label{sup:subsec:evaluation}
We evaluate our approach added to each of the baselines with a diverse range of attacks including strong adaptive attacks, summarized below. The first three are applied directly to the defended model.
\begin{itemize}
    \itemsep0em
    \item \textbf{AutoAttack} \cite{croce2020reliable} is an ensemble of attacks including two variants of the white-box AutoPGD (APGD) attack with cross-entropy (CE) loss and difference of logits ratio (DLR) loss, white-box fast adaptive boundary (FAB) attack \cite{croce2020minimally}, and a black-box Square attack \cite{andriushchenko2020square}. Note that AutoAttack evaluates the model following a sequence of attacks, and only the samples unsuccessfully perturbed by an attack are given to the next attack. 
    The hyperparameter settings for AutoAttack are followed from RobustBench \cite{croce2021robustbench}.    
    \item \textbf{Image-Wise Worst-Case} (IW-WC) robust accuracy is computed using an ensemble of 9 white-box, black-box, and adaptive attacks that we devised (extending the evaluation of \cite{croce2022evaluating}). The adaptive attacks specifically target our weakness of dependence on pseudo-labels. For each image, the attack is considered successful if any one of the 9 attacks is successful.
    We detail each attack included in the IW-WC robust accuracy computation below:
    \begin{enumerate}
    \item Standard \textbf{AutoAttack} \cite{croce2020reliable} applied to the defended model (\ie including the test-time defense). The hyperparameters are the same as in RobustBench \cite{croce2021robustbench}.
    \item \textbf{RayS} \cite{chen2020rays} is a black-box attack that searches for a perturbation that yields the lowest decision boundary radius. However, it is computationally very expensive and yet weaker than Square attack. We set the number of queries to $10000$ following \cite{croce2022evaluating,chen2020rays}. This is applied to the defended model.
    \item \textbf{APGD with EoT} \cite{athalye2018obfuscated}. Expectation over Transformation (EoT) averages the gradients over multiple iterations of each input to counter the effect of randomness in a defense. We use 20 iterations of EoT with APGD, which is more than $n_s=1$ used in our randomized smoothing. This is applied directly to the defended model.
\end{enumerate}
\end{itemize}
\begin{itemize}
    \item The following 6 attacks are computed on the base model and evaluated on the defended model (\ie these are transfer attacks). Also, while AutoAttack returns the unperturbed samples when the base model produces a misclassification, the following APGD attacks always return perturbed samples that misclassify or produce the highest loss. This is important because test-time defenses tend to misclassify samples closer to the boundary \cite{croce2022evaluating}.
    \begin{enumerate}
    \setcounter{enumi}{3}
    \item \textbf{Transfer-APGD-CE} attack uses the cross-entropy loss for 100 steps of APGD. 
    \item \textbf{Transfer-APGD-CW} attack uses Carlini-Wagner loss \cite{carlini2017towards} for 100 steps of APGD.
    \item \textbf{Transfer-Targeted-APGD-DLR} attack uses difference of logits ratio (DLR) loss \cite{croce2020reliable} for 100 steps of targeted APGD with 3 random restarts.
    \item \textbf{Transfer-AutoAttack} (same hyperparameters as 1).
    \item \textbf{Transfer-RayS} attack (same hyperparameters as 2).
    \item \textbf{Transfer-APGD+EoT} attack (same hyperparameters as 3).
    \end{enumerate}
\end{itemize}

\subsection{Miscellaneous details}
\label{sup:subsec:compute_details}

We implement our \textbf{IG-Defense} in PyTorch \cite{paszke2019pytorch} and use the official code releases from RobustBench \cite{croce2021robustbench}, RayS \cite{chen2020rays}, and Croce et al.\ \cite{croce2022evaluating} for our evaluations.
For base models, we use either RobustBench \cite{croce2021robustbench} implementations or the official base model authors' code releases.
For all experiments, we use a single Nvidia RTX 3090 GPU with 24 GB VRAM, and with an 8-core CPU and 16 GB RAM.

%% file: suppl/tabs/analysis_expt.tex
\begin{table}[t]
    \centering
    \setlength{\tabcolsep}{8pt}
    \caption{Analysis experiment results in table form (the average results, \ie last row, are illustrated in Fig.\ \ref{fig:analysis_expt} of the main paper).}
    \resizebox{1.0\linewidth}{!}{
\begin{tabular}{@{}cccccccc@{}}
\toprule
         & \multicolumn{4}{c}{\% change: successful attacks} & \multicolumn{3}{c}{\% change: unsuccessful attacks} \\
        \cmidrule(lr){2-5} \cmidrule(lr){6-8}
        Base Model & {\color[HTML]{674EA7}$\Delta_y$} & {\color[HTML]{c19b28}$\Delta_{\hat{y}^*}$} & {\color[HTML]{6aa84f}$\Delta_\text{rem-cls}$} & {\color[HTML]{dc4d4d}$\Delta_\text{unimp}$} & {\color[HTML]{674EA7}$\Delta_y$} & {\color[HTML]{6aa84f}$\Delta_\text{non-GT}$} & {\color[HTML]{dc4d4d}$\Delta_\text{unimp}$} \\
\midrule
GAT \cite{sriramanan2020guided}               & -17.84                                                                    & 15.83                                                                                    & -0.7087                          & -0.3593                                  & -51.00                            & -19.92                            & -16.1                                    \\
NuAT \cite{sriramanan2021towards}              & -62.67                                                                    & 56.85                                                                                    & -1.44                            & -2.01                                    & -84.77                         & -6.63                             & -25.06                                   \\
OAAT \cite{addepalli2022scaling}               & -21.49                                                                    & 20.31                                                                                    & -0.32                            & -0.015                                   & -52.37                         & -14.37                            & -21.35                                   \\
DAJAT \cite{addepalli2022efficient}              & -44.84                                                                    & 46.58                                                                                    & 0.86                             & 0.63                                     & -39.75                         & 5.06                              & -0.34                                    \\
TR-AWP \cite{wu2020adversarial}            & -19.80                                                                     & 23.34                                                                                    & 2.84                             & 0.54                                     & -13.36                         & 13.43                             & 4.48                                     \\
Average                    & -33.33                                                                   & 32.58                                                                                   & 0.246                          & -0.243                                 & -48.25                         & -4.48                            & -11.67                                  \\ \bottomrule
\end{tabular}
}
\label{supp:tab:analysis_expt_repeat}
\end{table}

%% file: suppl/3_analysis.tex
\section{Experiments}
\label{sup:sec:expts}

\input{suppl/tabs/test_time_comparison_ext}

\subsection{Extended comparisons}
\label{sup:subsec:extended_comp}

In Table \ref{sup:tab:comparison_test_time_ext}, we show the individual evaluations of the proposed image-wise worst-case (IW-WC) ensemble attack. While prior works show improvement over certain attacks, only our \textbf{IG-Defense} consistently improves over the base model. Further, we achieve significant gains in the IW-WC robust accuracy while maintaining the lowest inference time.

For existing test-time defenses as well as our \textbf{IG-Defense}, we find transfer attacks to be stronger than directly attacking the test-time defended model. 
Hence, we focus on evaluating against stronger attacks since improvements against weaker attacks are not considered fundamental gains in robustness \cite{croce2022evaluating}.
Specifically, we find that transfer of targeted APGD-DLR attack and transfer of APGD+EoT are the strongest attacks that break existing test-time defenses (HD, CAAA, SODEF). This is in-line with \cite{croce2022evaluating}, who show that HD breaks under the targeted APGD-DLR attack, SODEF breaks under an ensemble of transfer-APGD attacks, and CAAA breaks under any transfer attack. We showcase the robustness of \textbf{IG-Defense} against these strong adaptive attacks as well as against RayS, transfer-RayS, APGD+EoT, and transfer of APGD+EoT.

\begin{figure*}[h]
    \centering
    \includegraphics[width=1.0\linewidth]{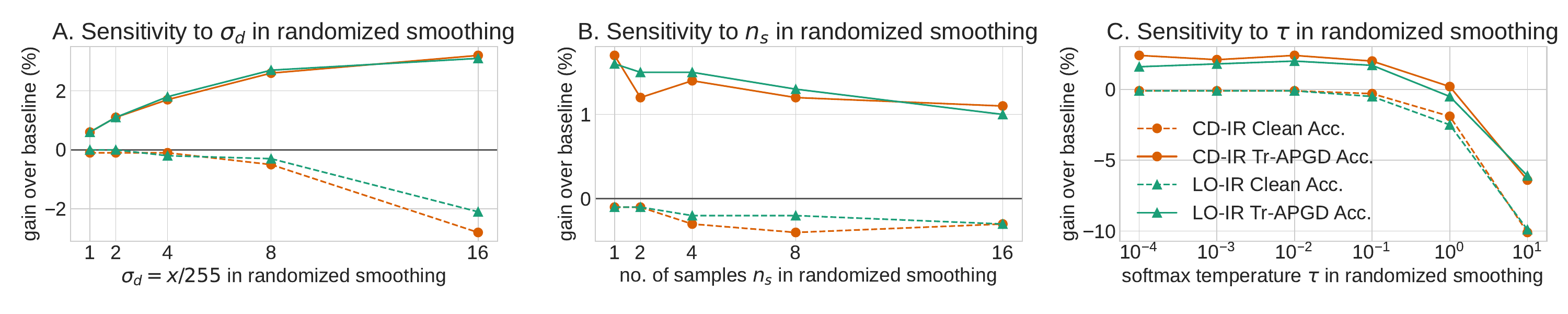}
    \caption{ 
        \textbf{Sensitivity to randomized smoothing hyperparameters.} The base model is DAJAT (ResNet18) \cite{addepalli2022efficient} for CIFAR10. The legend in \textbf{C.}\ applies to \textbf{A.} and \textbf{B.} as well.
    }
    \label{sup:fig:rs_sensitivity}
\end{figure*}

\subsection{Extended analysis}
\label{sup:subsec:sensitivity}

\textbf{Sensitivity analysis.}
Due to space constraints in the main paper, we present the sensitivity analyses for the randomized smoothing (RS) hyperparameters in Fig.\ \ref{sup:fig:rs_sensitivity}. Recall that RS adds $n_s$ number of noises sampled from a Gaussian distribution with zero mean and standard deviation $\sigma_d$. Formally,
\begin{equation}
    \hat{y} = 
    \sigma\left(
    \frac{1}{\tau} 
    \expectation_{v_i}
    [f(x+v_i)]
    \right); \;\; v_i \sim \mathcal{N}(0_d, \sigma_d^2 I_d)\; \forall \;i\in [n_s]
\end{equation}
where $0_d, I_d$ are tensors with shape same as $x$ but containing all zeros and ones respectively. $\sigma(\cdot)$ is the softmax function, and $\tau$ is the softmax temperature term that controls the sharpness of the softmax distribution. Overall, we have three hyperparameters, $n_s, \sigma_d$, and $\tau$.

In Fig.\ \ref{sup:fig:rs_sensitivity}\red{A}, we analyze the sensitivity to $\sigma_d$ by varying it from $1/255$ to $16/255$ while $n_s\!=\!1$. While the robust accuracy gain increases with $\sigma_d$, the clean accuracy drop also increases for both \texttt{CD-IR} and \texttt{LO-IR}. Hence, we choose $\sigma_d\!=\!\frac{\epsilon}{2}$ (\ie $4/255$ in Fig.\ \ref{sup:fig:rs_sensitivity}\red{A} for CIFAR10) where the clean accuracy drop is minimal with significant robustness gains.

In Fig.\ \ref{sup:fig:rs_sensitivity}\red{B}, we analyze the sensitivity to number of noises sampled $n_s$ by varying it from $1$ to $16$ while $\sigma_d$ is fixed to $4/255$. There is a consistent robustness gain across all $n_s$. But the clean accuracy drop as well as inference time is lower for a lower $n_s$. Hence, we choose $n_s=1$.

In Fig.\ \ref{sup:fig:rs_sensitivity}\red{C}, we analyze the sensitivity to the softmax temperature $\tau$ by varying it from $10^{-4}$ to $10$. We find that a higher softmax temperature leads to a drop in both clean and robust accuracy since the initial pseudo-label may be wrong if the distribution is not sharp enough. Between $10^{-1}$ to $10^{-4}$, we get almost the same performance gains and both \texttt{CD-IR} and \texttt{LO-IR} are fairly insensitive to $\tau$.

\begin{figure}[h]
    \centering
    \includegraphics[width=\linewidth]{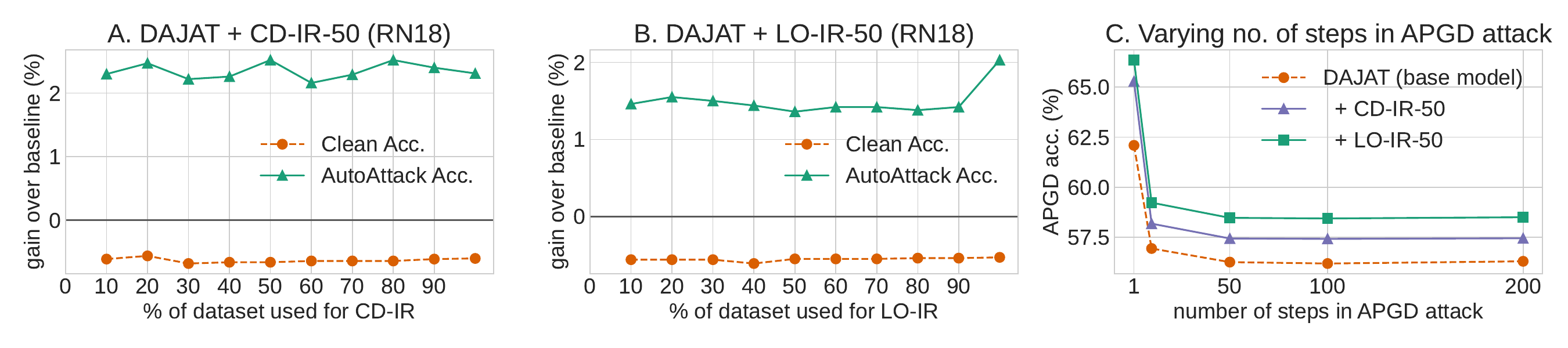}
    \caption{ 
        \newblue{\textbf{A, B.} Sensitivity to amount of dataset used for our importance ranking methods, \texttt{CD-IR} and \texttt{LO-IR}.}
        \textbf{C.} Varying the number of steps in APGD attack \cite{croce2020reliable} on our \textbf{IG-Defense} with DAJAT \cite{addepalli2022efficient} (ResNet18) for CIFAR10.
    }
    \label{sup:fig:varying_steps}
\end{figure}

\newblue{
A potential limitation of \textbf{IG-Defense} is that our importance ranking methods require access to probing data, which is a general requirement to use neuron interpretability tools. In our experiments, we use the training data as the probing data. Hence, in Fig.\ \ref{sup:fig:varying_steps}\red{A} and Fig.\ \ref{sup:fig:varying_steps}\red{B}, we analyze the sensitivity to the amount of data required by our importance ranking methods, \texttt{CD-IR} and \texttt{LO-IR}. We find that even 10\% of the probing data is sufficient for both importance ranking methods to yield robustness gains.
}

\noindent
\textbf{Sanity checks for gradient masking.}

In Table \ref{sup:tab:comparison_test_time}, we evaluate with black-box Square attack and stronger AutoAttack on CIFAR10 with the ResNet18 base model from DAJAT \cite{addepalli2022efficient}. Specifically, we add a stronger AutoAttack evaluation where APGD uses 200 steps with 10 random restarts (RR) instead of 100 steps and 1 RR, targeted FAB attack also uses 10 RR instead of 1 RR, and Square attack uses 5 RR instead of 1 RR. Compared to the standard AutoAttack, the accuracy reduction for stronger AutoAttack is only marginal for all methods except SODEF (which has $\sim$9\% drop). This shows that our methods are strong defenses irrespective of attack strength although SODEF is more vulnerable. Further, the black-box Square attack is less effective compared to the white-box AutoAttack indicating that gradients are useful, \ie not masked. However, gradient masking exists in CAAA and SODEF since the black-box Square (Table \ref{sup:tab:comparison_test_time}) and RayS attacks (Table \ref{sup:tab:comparison_test_time_ext}) respectively are stronger than the white-box AutoAttack.

\input{suppl/tabs/test_time_comparisons}

\begin{figure*}[h]
    \centering
    \includegraphics[width=1.0\linewidth]{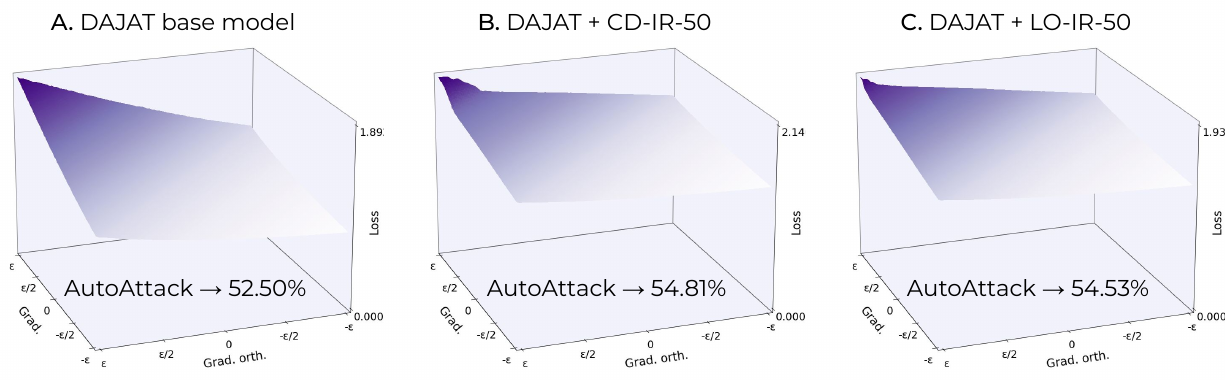}
    \caption{ 
        Comparison of loss surfaces for DAJAT \cite{addepalli2022efficient} and DAJAT with our \textbf{IG-Defense} at test-time.
    }
    \label{sup:fig:loss_surface}
\end{figure*}

In Fig.\ \ref{sup:fig:loss_surface}, we compare the loss surfaces or landscapes for the DAJAT \cite{addepalli2022efficient} ResNet18 base model on CIFAR10 and with our proposed \textbf{IG-Defense}. 
Specifically, the loss is computed for perturbed examples $x' = x + \alpha g + \beta g^\perp$ where $\alpha, \beta$ are varied to cover the entire $l_\infty$-norm $\epsilon$-ball, $g$ is the sign of gradient and $g^\perp$ is a direction orthogonal to $g$.
We observe that loss surfaces are smooth for both \texttt{CD-IR} and \texttt{LO-IR}, indicating the absence of gradient masking since even a brute-force search in the $\epsilon$-ball does not yield a higher loss than along the gradient direction.

In Fig.\ \ref{sup:fig:varying_steps}\red{C}, we experiment with single-step (\ie FGSM) and multi-step APGD attacks. We see consistent gains in robustness irrespective of number of attack steps. Multi-step attacks are stronger than single-step attack, indicating that gradients are reliable. Further, as number of steps are increased, robust accuracy saturates similar to the base model, indicating that gradients are as useful as in the base model.

\input{suppl/tabs/varying_eps}

In Table \ref{sup:tab:varying_eps}, we vary the $\ell_\infty$-bound $\epsilon$ of AutoAttack from $4/255$ to $32/255$. As $\epsilon$ is increased, the base model and our defense robustness both can be reduced to almost zero. This again indicates that gradients are reliable.

\input{suppl/tabs/class_disparity}

\noindent
\newblue{
\textbf{Classwise disparity in adversarially-trained models.}}

\newblue{
Prior works \cite{xu2021robust, wei2023cfa} have shown that adversarial training leads to significant disparity in class-wise robustness, which is also known as the robust fairness issue. We investigate this by evaluating the class-wise robust accuracies and report the best and worst class accuracies in Table \ref{sup:tab:class_disparity}.
We observe similar (though relatively less severe) disparity results. It can also be seen that our test-time defenses yield uniform robustness gains across classes.
}

\subsection{Efficiency analysis}
\label{sup:subsec:efficiency}

We present the efficiency analysis for our proposed importance ranking methods \texttt{CD-IR} and \texttt{LO-IR} in Table \ref{sup:tab:efficiency}.
For \texttt{CD-IR}, most of the operations are vectorized and the training dataset is passed through the CLIP image encoder and the base model once while saving the CLIP embeddings and base model activations. Then, the similarity can be computed easily with the saved activations and we observe that \texttt{CD-IR} can be performed very quickly ($\sim$2 minutes).

For \texttt{LO-IR}, we need to compute the average logit change when masking each neuron separately. Given that each computation is independent and inference requires a very low amount of GPU memory (as it does not require backpropagation), we parallelize this into 8 threads. With this, we can perform \texttt{LO-IR} for 512 neurons in $\sim$28 minutes.

\input{suppl/tabs/efficiency_analysis}

\newblue{A potential concern about compute cost could be due to hyperparameter search required for the number of retained neurons $k$. However, we observe that the compute cost to get $k$ is actually not very high. This is because our importance ranking is independent of $k$, \ie it only needs to be done once. Then, we can use binary search to find $k$, which requires evaluating clean accuracy ($\sim$2 mins per $k$). This is much cheaper than computing the robust accuracy ($\sim$6 hrs per $k$). 
Because, as a heuristic, we usually find a robustness gain when the clean accuracy is close to that of the base model.}

%% file: suppl/tabs/test_time_comparison_ext.tex
\begin{table}[t]
    \centering
    \setlength{\tabcolsep}{5pt}
    \caption{Extended comparison with state-of-the-art test-time defenses on CIFAR10 (extending Table \ref{tab:comparison_test_time}, main paper). The number in \darkgreen{(green)} shows the worst-case robustness gain over the base model. The first row of each block is the base model without test-time defense, AA indicates AutoAttack \cite{croce2020reliable}, Tr indicates Transfer attack.
    The \textbf{last column (IW-WC)} indicates image-wise worst-case robust accuracy over the 9 attacks.
    }
    \resizebox{\linewidth}{!}{
\begin{tabular}{clccccccccccl}
\toprule
\multirow{3}{*}{\begin{tabular}[c]{@{}l@{}}Base\\ Model\end{tabular}} &
  \multirow{3}{*}{\begin{tabular}[c]{@{}l@{}}Test-\\Time\\ Defense\end{tabular}} &
  \multirow{3}{*}{\begin{tabular}[c]{@{}c@{}}Clean\\ Acc.\end{tabular}} &
  \multicolumn{3}{c}{Attacks on test-time def.} &
  \multicolumn{6}{c}{Attacks transferred from base model} &
  \multirow{3}{*}{\begin{tabular}[c]{@{}c@{}}\textbf{Robust} \\ \textbf{Acc.}\\ \textbf{(IW-WC)}\end{tabular}} \\ \cmidrule(lr){4-6}\cmidrule(lr){7-12}
 &
   &
   &
  AA &
  RayS &
  \begin{tabular}[c]{@{}c@{}}APGD\\ +EoT\end{tabular} &
  \begin{tabular}[c]{@{}c@{}}Tr-APGD\\ -CE\end{tabular} &
  \begin{tabular}[c]{@{}c@{}}Tr-APGD\\ -CW\end{tabular} &
  \begin{tabular}[c]{@{}c@{}}Tr-APGD \\ -tgt-DLR\end{tabular} &
  Tr-AA &
  \begin{tabular}[c]{@{}c@{}}Tr-\\ RayS\end{tabular} &
  \begin{tabular}[c]{@{}c@{}}Tr-APGD\\ +EoT\end{tabular} &
   \\ \midrule 
\multirow{6}{*}{\begin{tabular}[c]{@{}c@{}}OAAT \\ (RN18)\end{tabular}} &
  None &
  80.23 &
  51.10 &
  56.74 &
  51.61 &
  55.72 &
  51.73 &
  51.10 &
  51.10 &
  56.74 &
  51.61 &
  51.01 \\
 &
  HD &
  79.89 &
  62.45 &
  73.24 &
  63.20 &
  58.86 &
  55.12 &
  50.76 &
  59.19 &
  71.77 &
  59.46 &
  50.68 {\smaller\lightred{(-0.33)}} \\
 &
  SODEF &
  80.23 &
  59.05 &
  74.55 &
  62.55 &
  55.09 &
  52.94 &
  57.80 &
  63.09 &
  74.33 &
  63.45 &
  50.67 {\smaller\lightred{(-0.34)}} \\
 &
  CAAA &
  80.23 &
  69.45 &
  56.93 &
  71.49 &
  55.72 &
  51.73 &
  51.12 &
  51.04 &
  56.77 &
  51.61 &
  51.00 {\smaller\lightred{(-0.01)}} \\
 &
 \cellcolor{gray!10}\texttt{CD-IR}-50 &
  \cellcolor{gray!10}79.85 &
  \cellcolor{gray!10}54.79 &
  \cellcolor{gray!10}79.80 &
  \cellcolor{gray!10}55.86 &
  \cellcolor{gray!10}56.69 &
  \cellcolor{gray!10}52.58 &
  \cellcolor{gray!10}53.28 &
  \cellcolor{gray!10}54.61 &
  \cellcolor{gray!10}68.80 &
  \cellcolor{gray!10}54.95 &
  \cellcolor{gray!10}\textbf{52.35 {\smaller\darkgreen{(+1.34)}}} \\
 &
 \cellcolor{gray!10}\texttt{LO-IR}-50 &
  \cellcolor{gray!10}79.90 &
  \cellcolor{gray!10}54.90 &
  \cellcolor{gray!10}79.77 &
  \cellcolor{gray!10}56.23 &
  \cellcolor{gray!10}56.59 &
  \cellcolor{gray!10}52.63 &
  \cellcolor{gray!10}53.28 &
  \cellcolor{gray!10}54.60 &
  \cellcolor{gray!10}68.93 &
  \cellcolor{gray!10}54.93 &
  \cellcolor{gray!10}52.30 {\smaller\darkgreen{(+1.29)}} \\
  \midrule
\multirow{6}{*}{\begin{tabular}[c]{@{}c@{}}DAJAT \\ (RN18)\end{tabular}} &
  None &
  85.71 &
  52.50 &
  59.94 &
  52.95 &
  56.24 &
  53.26 &
  52.53 &
  52.50 &
  59.94 &
  52.95 &
  52.45 \\
 &
  HD &
  84.53 &
  64.92 &
  78.22 &
  66.38 &
  60.46 &
  57.20 &
  52.67 &
  61.24 &
  75.73 &
  61.44 &
  52.55 {\smaller\darkgreen{(+0.10)}} \\
 &
  SODEF &
  84.86 &
  64.11 &
  82.06 &
  65.09 &
  56.27 &
  54.60 &
  57.42 &
  61.38 &
  76.41 &
  61.64 &
  52.95 {\smaller\darkgreen{(+0.50)}} \\
 &
  CAAA &
  85.71 &
  75.51 &
  60.14 &
  76.64 &
  56.24 &
  53.26 &
  52.53 &
  52.51 &
  59.96 &
  52.95 &
  52.45 {\smaller\lightred{(+0.00)}} \\
 &
 \cellcolor{gray!10}\texttt{CD-IR}-50 &
  \cellcolor{gray!10}85.11 &
  \cellcolor{gray!10}54.81 &
  \cellcolor{gray!10}85.00 &
  \cellcolor{gray!10}55.59 &
  \cellcolor{gray!10}57.03 &
  \cellcolor{gray!10}54.11 &
  \cellcolor{gray!10}54.20 &
  \cellcolor{gray!10}55.68 &
  \cellcolor{gray!10}72.73 &
  \cellcolor{gray!10}56.04 &
  \cellcolor{gray!10}\textbf{53.36 {\smaller\darkgreen{(+0.91)}}} \\
 &
 \cellcolor{gray!10}\texttt{LO-IR}-50 &
  \cellcolor{gray!10}85.18 &
  \cellcolor{gray!10}54.53 &
  \cellcolor{gray!10}85.08 &
  \cellcolor{gray!10}55.99 &
  \cellcolor{gray!10}57.00 &
  \cellcolor{gray!10}54.17 &
  \cellcolor{gray!10}54.20 &
  \cellcolor{gray!10}55.67 &
  \cellcolor{gray!10}72.14 &
  \cellcolor{gray!10}56.02 &
  \cellcolor{gray!10}53.34 {\smaller\darkgreen{(+0.89)}} \\
  \midrule
\multirow{6}{*}{\begin{tabular}[c]{@{}c@{}}TR-AWP \\ (WRN)\end{tabular}} &
  None &
  85.36 &
  56.17 &
  61.68 &
  56.42 &
  58.83 &
  56.76 &
  56.25 &
  56.17 &
  61.68 &
  56.42 &
  56.12 \\
 &
  HD &
  84.78 &
  68.61 &
  78.84 &
  69.51 &
  61.70 &
  60.15 &
  55.79 &
  63.71 &
  76.56 &
  63.87 &
  55.72 {\smaller\lightred{(-0.40)}} \\
 &
  SODEF &
  85.25 &
  64.46 &
  79.41 &
  66.57 &
  57.84 &
  57.66 &
  60.30 &
  63.81 &
  75.85 &
  63.92 &
  56.05 {\smaller\lightred{(-0.07)}} \\
 &
  CAAA &
  85.35 &
  75.74 &
  61.75 &
  77.47 &
  58.83 &
  56.75 &
  56.25 &
  56.20 &
  61.76 &
  56.43 &
  56.11 {\smaller\lightred{(-0.01)}} \\
 &
 \cellcolor{gray!10}\texttt{CD-IR}-50 &
  \cellcolor{gray!10}84.98 &
  \cellcolor{gray!10}60.26 &
  \cellcolor{gray!10}83.84 &
  \cellcolor{gray!10}61.41 &
  \cellcolor{gray!10}59.73 &
  \cellcolor{gray!10}58.02 &
  \cellcolor{gray!10}57.53 &
  \cellcolor{gray!10}59.43 &
  \cellcolor{gray!10}72.74 &
  \cellcolor{gray!10}59.60 &
  \cellcolor{gray!10}57.46 {\smaller\darkgreen{(+1.34)}} \\
 &
 \cellcolor{gray!10}\texttt{LO-IR}-50 &
  \cellcolor{gray!10}85.17 &
  \cellcolor{gray!10}59.61 &
  \cellcolor{gray!10}85.03 &
  \cellcolor{gray!10}60.47 &
  \cellcolor{gray!10}59.07 &
  \cellcolor{gray!10}57.32 &
  \cellcolor{gray!10}57.58 &
  \cellcolor{gray!10}59.02 &
  \cellcolor{gray!10}72.98 &
  \cellcolor{gray!10}59.17 &
  \cellcolor{gray!10}\textbf{57.88 {\smaller\darkgreen{(+1.76)}}} \\ 
\bottomrule 
\end{tabular}
    }
    \label{sup:tab:comparison_test_time_ext}
\end{table}

%% file: suppl/tabs/test_time_comparisons.tex
\begin{table}[h]
    \centering
    \setlength{\tabcolsep}{5pt}
    \caption{Sanity check for gradient masking on CIFAR10. The number in \darkgreen{(green)} shows the worst-case robustness gain over the base model. The first row is the base model without test-time defense, AA indicates AutoAttack \cite{croce2020reliable}.}
    \resizebox{\linewidth}{!}{
        \begin{tabular}{llccccl}
            \toprule
            Base Model & Test-Time Defense & \begin{tabular}[c]{@{}c@{}}Clean\\ Acc.\end{tabular} & \begin{tabular}[c]{@{}c@{}}Robust Acc.\\ (AA)\end{tabular} & \begin{tabular}[c]{@{}c@{}}Robust Acc.\\ (Stronger AA)\end{tabular} & \begin{tabular}[c]{@{}c@{}}Robust Acc.\\ (Square)\end{tabular} & \begin{tabular}[c]{@{}c@{}}Robust Acc.\\ (IW-WC)\end{tabular} \\
            \midrule
            \multirow{5}{*}{\makecell{DAJAT \cite{addepalli2022efficient} \\ RN18}} & None & 85.71 & 52.50 & 52.45 & 59.23 & 52.45 \\
            \cmidrule{3-7}
             & CAAA \cite{alfarra2022combating} & 85.71 & 75.51 & 74.61 & 78.51 & 52.45 \darkgreen{\small(+0.00)} \\
             & SODEF \cite{kang2021stable} & 84.44 & 64.11 & 55.09 & 59.54 & 52.95 \darkgreen{\small(+0.50)} \\
             & \texttt{CD-IR}-50 (\textit{Ours}) & 85.11 & 54.81 & 53.87 & 61.76 & \textbf{53.36 \darkgreen{\small(+0.91)}} \\
             & \texttt{LO-IR}-50 (\textit{Ours}) & 85.18 & 54.53 & 53.93 & 62.22 & {53.34 \darkgreen{\small(+0.89)}} \\
            \bottomrule
        \end{tabular}%
    }
    \label{sup:tab:comparison_test_time}
\end{table}

%% file: suppl/tabs/varying_eps.tex
\begin{table}[h]
    \centering
    \setlength{\tabcolsep}{12pt}
    \caption{Varying $\ell_\infty$-bound $\epsilon$ of AutoAttack for ResNet18 with CIFAR10. We omit clean accuracy since it does not change with $\epsilon$.}
    \resizebox{0.7\linewidth}{!}{
        \begin{tabular}{ccrr}
            \toprule
            $\epsilon$ & DAJAT \cite{addepalli2022efficient} & +\texttt{CD-IR}-50 & +\texttt{LO-IR}-50 \\
            \midrule
            $4/255$ & 71.10 & 71.63 \darkgreen{\small(+0.53)} & 71.44 \darkgreen{\small(+0.34)} \\
            $8/255$ & 52.50 & 54.81 \darkgreen{\small(+2.31)} & 54.53 \darkgreen{\small(+2.03)} \\
            $16/255$ & 17.93 & 20.00 \darkgreen{\small(+2.07)} & 19.75 \darkgreen{\small(+1.82)} \\
            $32/255$ & $\;\;$0.52 & 0.71 \darkgreen{\small(+0.19)} & 0.72 \darkgreen{\small(+0.20)} \\
            \bottomrule
        \end{tabular}%
    }
    \label{sup:tab:varying_eps}
\end{table}

%% file: suppl/tabs/class_disparity.tex
\begin{table}[h]
    \centering
    \setlength{\tabcolsep}{7pt}
    \caption{\newblue{Analysis of class-wise accuracies for best and worst classes for the DAJAT \cite{addepalli2022efficient} ResNet18 base model on CIFAR10}}
    \resizebox{\linewidth}{!}{
    \begin{tabular}{lcccccc}
        \toprule
         & \multicolumn{2}{c}{Classwise Average} & \multicolumn{2}{c}{Best-Class} & \multicolumn{2}{c}{Worst-Class} \\
        \cmidrule(lr){2-3} \cmidrule(lr){4-5} \cmidrule(lr){6-7}
         & \makecell{Clean \\ Acc.} & \makecell{Rob. Acc. \\ (AA)} & \makecell{Clean \\ Acc.} & \makecell{Rob. Acc. \\ (AA)} & \makecell{Clean \\ Acc.} & \makecell{Rob. Acc. \\ (AA)} \\
        \midrule
        No test-time def.\ & \textbf{85.71} & 52.50 & \textbf{94.4} & 74.7 & \textbf{72.3} & 25.7 \\
        \texttt{CD-IR}-50 {\small \textit{(Ours)}} & 85.11 & \textbf{54.81} & 94.3 & 76.5 & 71.8 & \textbf{28.8} \\
        \texttt{LO-IR}-50 {\small \textit{(Ours)}} & 85.18 & 54.53 & 94.2 & \textbf{76.6} & 72.1 & 28.5 \\
        \bottomrule
    \end{tabular}%
    }
    \label{sup:tab:class_disparity}
\end{table}

%% file: suppl/tabs/efficiency_analysis.tex
\begin{table}[h]
    \centering
    \setlength{\tabcolsep}{18pt}
    \caption{Efficiency analysis of our proposed importance ranking methods \texttt{CD-IR} and \texttt{LO-IR} for layer4 (with 512 neurons) of a ResNet18 base model and CIFAR10. Both experiments performed with a single RTX 3090 GPU with 24 GB VRAM, 8-core CPU, and 16 GB RAM.}
    \resizebox{0.4\linewidth}{!}{
        \begin{tabular}{ccrr}
            \toprule
            Method & Time taken \\
            \midrule
            \texttt{LO-IR} & $\sim$28 mins. \\
            \texttt{CD-IR} & $\sim$2 mins. \\
            \bottomrule
        \end{tabular}%
    }
    \label{sup:tab:efficiency}
\end{table}